\pdfoutput=1

\documentclass[11pt]{article}

\usepackage{acl}

\usepackage{times}
\usepackage{latexsym}
\usepackage{tabularx}
\usepackage{graphicx}
\usepackage{amsmath}
\usepackage{multirow}
\usepackage{pifont}

\usepackage[T1]{fontenc}

\usepackage[utf8]{inputenc}

\usepackage{microtype}

\usepackage{inconsolata}

\title{Rule or Story, Which is a Better Commonsense Expression for Talking with Large Language Models? }

\author{Ning Bian$^{\rm 1,2}$, Xianpei Han$^{\rm 1,2,}$\thanks{Corresponding Author}, Hongyu Lin$^{\rm 2}$, Yaojie Lu$^{\rm 2}$, Ben He$^{\rm 1,2}$, Le Sun$^{\rm 1,2}$\\
  $^{\rm 1}$University of Chinese Academy of Sciences, Beijing, China \\
  $^{\rm 2}$Institute of Software, Chinese Academy of Sciences, Beijing, China \\
  \tt bianning21@mails.ucas.ac.cn \\
  \tt \{xianpei,hongyu,yaojie,sunle\}@iscas.ac.cn\ \ \tt benhe@ucas.ac.cn\\}

\begin{document}
\maketitle

\begin{abstract}
Building machines with commonsense has been a longstanding challenge in NLP due to the reporting bias of commonsense rules and the exposure bias of rule-based commonsense reasoning. In contrast, humans convey and pass down commonsense implicitly through stories. This paper investigates the inherent commonsense ability of large language models (LLMs) expressed through storytelling. We systematically investigate and compare stories and rules for retrieving and leveraging commonsense in LLMs. Experimental results on 28 commonsense QA datasets show that stories outperform rules as the expression for retrieving commonsense from LLMs, exhibiting higher generation confidence and commonsense accuracy. Moreover, stories are the more effective commonsense expression for answering questions regarding daily events, while rules are more effective for scientific questions. This aligns with the reporting bias of commonsense in text corpora. We further show that the correctness and relevance of commonsense stories can be further improved via iterative self-supervised fine-tuning. These findings emphasize the importance of using appropriate language to express, retrieve, and leverage commonsense for LLMs, highlighting a promising direction for better exploiting their commonsense abilities.
\end{abstract}

\section{Introduction}

Building machines with commonsense has been a longstanding goal in AI and NLP \cite{McCarthy_Programs59, brachman2023machines}. Despite advancements in large language models (LLMs), incorporating commonsense knowledge in these models remains a significant challenge \cite{ismayilzada2023crow, bian2023chatgpt, li-etal-2022-systematic}, due to the reporting bias of commonsense knowledge and the exposure bias of commonsense reasoning \cite{gordon2013reporting, shwartz-choi-2020-neural}.
The reporting bias arises because many aspects of commonsense are rarely stated explicitly in language. For example, ``\textit{A person is late}'' may appear more frequently than ``\textit{A person arrives on time}'' in text corpora \cite{gordon2013reporting}. Furthermore, commonsense rules are often left implicit and omitted in human language reasoning, leading to exposure bias. For example, the commonsense rule ``\textit{humans need air to breathe}'' is usually ignored in cases like ``\textit{The room was getting too stuffy, and I opened the windows}'' as it is commonly known.

\begin{figure}[!t]
\setlength{\belowcaptionskip}{-0.4cm}
  \centering
  \includegraphics[width=\columnwidth]{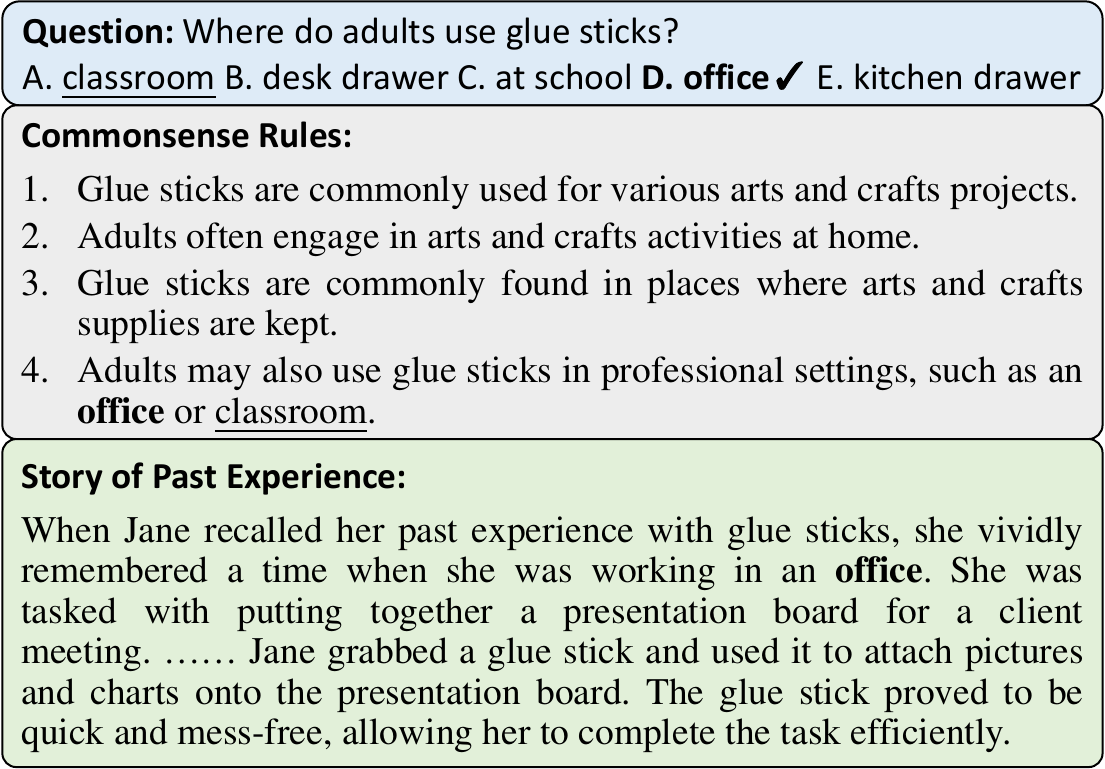}
  \caption{Comparison between rules and a story written by ChatGPT. The rules only provide useful knowledge until the 4$^{th}$ rule and also include an incorrect answer option, ``classroom''. The story presents a detailed scenario where an adult uses glue sticks in an office.}
  \label{f0}
\end{figure}

To enhance the commonsense ability of NLP models, current studies usually express commonsense as rules. For instance,  commonsense rules structured as knowledge graphs of concepts and events \cite{ilievski2021cskg, hwang2021comet, sap2019atomic, speer2017conceptnet} are incorporated to support rule-based logical reasoning \cite{zhang2023heuristic, wang2023car, wang-etal-2023-cat, wang-etal-2023-dynamic, wang-etal-2023-cola, liu2023mind}. Recently, as studies reveal that LLMs like GPT-3 \cite{brown2020language} and ChatGPT \cite{chatgpt} have already learned abundant commonsense \cite{shwartz-etal-2020-unsupervised}, there is a current trend to extract commonsense knowledge from the models' memory, also expressed as rules like in Figure \ref{f0}, and enhance LLMs by reintegrating this knowledge into the models \cite{liu2023crystal, yao2023knowledge, liu2022generated, west2021symbolic}. 

However, commonsense is more than just rules \cite{brachman2022toward, mostafazadeh2020glucose}. Humans acquire commonsense by recognizing prototypical patterns, extracting memories of similar past experiences, and contrasting them with the current novel situation to make decisions, as supported by psychological studies \cite{schacter2007cognitive, gary2004power, tulving2002episodic, schank1983dynamic, reason:SchAbe77a}. Our commonsense is often conveyed and passed down through stories such as myths and fairy tales \cite{cassirer1946language}, with only a limited portion expressed in rules. Renowned AI theorist and cognitive psychologist Roger Schank argues in his book ``\textit{Tell Me a Story: Narrative and Intelligence}'' that ``\textit{knowledge is stories}'' \cite{schank1995tell}. He emphasizes that humans struggle to learn and remember abstract rules derived from past experiences but can more easily remember a good story, because ``\textit{stories give life to past experience}''.


As a result, human-written text corpora mainly convey commonsense through stories, with limited instances of explicit rules and logical reasoning. In this way, models trained on these corpora acquire commonsense and reasoning abilities implicitly. Studies show that LLMs exhibit a strong storytelling ability, generating narratives that adhere to real-world logic \cite{bhandari2023trustworthiness, eldan2023tinystories, wen-etal-2023-grove, jiayang-etal-2023-storyanalogy}. However, these models may not effectively learn commonsense rules and explicit reasoning through mimicking human behaviors, as shown by recent studies \cite{bian2023chatgpt, li-etal-2022-systematic}. 

These observations lead to a critical question: Which is the better commonsense expression for talking with LLMs—rule or story? Specifically, this paper aims to answer the following two questions:
(1) Which expression is more effective for retrieving commonsense from the memory of LLMs? (2) Which expression is more suitable for LLMs to leverage commonsense in solving problems?

To answer the questions, we systematically compare stories and rules as commonsense expressions for talking with LLMs. We use a total of 28 commonsense QA datasets for experiments.
For the first question, we instruct LLMs to generate stories and rules based on commonsense questions, as shown in Figure \ref{f0}. We compare the confidence and the accuracy of commonsense generation using stories and rules, showing that LLMs are more confident and more accurate at retrieving commonsense as stories than as rules. 
For the second question, we compare the confidence of generating the correct answers with stories or rules as contexts, showing that LLMs can more confidently leverage stories than rules for reasoning. The QA accuracy results further demonstrate that the story is a more effective commonsense expression for answering questions regarding daily events, while the rule is more effective for scientific commonsense QA. This phenomenon aligns with the reporting bias of commonsense in the text corpora. Moreover, stories and rules can complement each other, i.e., combining them can further enhance the answer accuracy.

In-depth analyses reveal two main issues in generating commonsense stories: commonsense hallucination and semantic drifting. To address these problems, we propose an iterative self-supervised fine-tuning (self-SFT) method. We ask the model to generate stories given the training set of 8 datasets and design a scoring method to rank the stories based on their consistency with commonsense and similarity with the question. We filter the stories based on the scores and use them to fine-tune the model. The tuned model is then used to generate stories in the next iteration. 
Experimental results show that the self-SFT method leads to further accuracy improvements, highlighting the potential for LLMs to self-improve their commonsense abilities.

The main contributions of this paper are:

1. We systematically investigate and compare the effects of using stories and rules as commonsense expressions for retrieving and leveraging commonsense in LLMs. To our best knowledge, this is the first study to investigate the effects of specific commonsense expressions in LLMs.

2. We show that the story is a more effective expression for retrieving commonsense from LLMs and for leveraging commonsense in answering questions regarding daily events.

3. We identify two main issues that hinder commonsense story generation: commonsense hallucination and semantic drifting, and propose an iterative self-SFT method to improve the accuracy and relevance of stories generated by LLMs.

\section{Background}

\paragraph{Commonsense QA.}

Answering commonsense questions has become one of the standard tasks for evaluating LLMs \cite{srivastava2022beyond}. In this paper, we use 28 commonsense QA datasets for experiments, covering different domains of commonsense. These datasets are CommonsenseQA \cite{talmor-etal-2019-commonsenseqa}, OpenBookQA \cite{mihaylov-etal-2018-suit}, Winograd Schema Challenge (WSC) \cite{levesque2012winograd}, PIQA \cite{bisk2020piqa}, SocialIQA \cite{sap-etal-2019-social}, ARC (easy and challenge set) \cite{clark2018think}, QASC \cite{khot2020qasc}, HellaSWAG (ActivitiNet and WikiHow set) \cite{zellers-etal-2019-hellaswag}, NumerSense \cite{lin-etal-2020-birds}, AI2 Science Questions (AI2Sci, elementary and middle school set) \cite{AI2Sci}, CommonsenseQA 2.0 \cite{talmor2021commonsenseqa}, SWAG \cite{zellers-etal-2018-swag}, WinoGrande \cite{sakaguchi2021winogrande}, Com2Sense \cite{singh-etal-2021-com2sense}, SciQ \cite{welbl-etal-2017-crowdsourcing}, QuaRel \cite{tafjord2019quarel}, QuaRTz \cite{tafjord-etal-2019-quartz}, CycIC, ComVE (Task A) \cite{wang-etal-2019-make}, COPA \cite{roemmele2011choice}, PROST \cite{aroca-ouellette-etal-2021-prost}, CODAH \cite{chen-etal-2019-codah}, Story Cloze Test (SCT) \cite{mostafazadeh-etal-2016-corpus}, $\alpha$NLI \cite{bhagavatula2019abductive}, and WinoVenti \cite{do-pavlick-2021-rotten}. We use their development set for evaluation.

\textbf{Commonsense knowledge and knowledge-augmented reasoning.}
In commonsense research, there is a growing consensus that integrating knowledge can improve the commonsense ability of NLP models \cite{bian2021benchmarking}. Typically, commonsense knowledge is expressed by concise and clear rules as either triples of <head, relation, tail> like in ConceptNet \cite{speer2017conceptnet}, or simple sentences like in Open Mind Common Sense \cite{10.5555/646748.701499}. Recently, researchers turn to retrieving commonsense knowledge from pre-trained LLMs like GPT-3, assuming LLMs have already learned abundant commonsense from large-scale human-written text corpora \cite{wang2023boosting, chen-etal-2023-distinguish, liu2023crystal, li2023guiding, yao2023knowledge, wang2023scott, zhou-etal-2022-think, wang2022pinto, wei2022chain, liu-etal-2022-rainier, liu2022generated, gu-etal-2022-dream, yu2022generate,  paranjape-etal-2021-prompting, west2021symbolic, bosselut2021dynamic, shwartz-etal-2020-unsupervised, latcinnik2020explaining, rajani-etal-2019-explain}. They instruct LLMs with prompts and examples to generate commonsense rules as concise sentences. In this paper, we follow their assumption and compare stories and rules as the commonsense expression for retrieving commonsense knowledge from LLMs.

There have been numerous works that inject commonsense rules into NLP models to improve commonsense reasoning, either by pre-training on knowledge bases \cite{ma2021knowledge, chang-etal-2020-incorporating, mitra2019additional, zhong2019improving}, or incorporating knowledge rules in the input of language models \cite{shi2023qadynamics, wang2023car, wang-etal-2023-cola, lal2022using, bian2021benchmarking}. There is also a chain of works that use graph-based reasoning for inference \cite{wang-etal-2023-dynamic, yasunaga-etal-2021-qa, lv2020graph, lin-etal-2019-kagnet}. In this paper, we exploit the commonsense knowledge in the memory of LLMs as stories or rules to support commonsense reasoning.

\textbf{Story generation by LLMs.}
Generating fluent and coherent stories using NLP models has been a crucial component of computational creativity. 
Recent studies have found that LLMs can perform well on story generation \cite{eldan2023tinystories, wen-etal-2023-grove, peng2021inferring}. \citet{bhandari2023trustworthiness} compare stories generated by OPT \cite{zhang2022opt}, LLaMA \cite{touvron2023llama}, and Alpaca \cite{alpaca} with human-written stories, showing that these two kinds of stories exhibit a remarkable similarity in terms of readability and topics. The LLM-generated stories are even more accessible than traditional children's stories. \citet{xie2023next} compared GPT-3 with story generation models before LLMs, demonstrating that LLMs generate stories of significantly high quality, even comparable with human authors.
In this paper, we analyze and exploit the inherent commonsense embedded in the storytelling ability of LLMs.

\textbf{Large language models.}
This study focuses on three LLMs: ChatGPT \cite{chatgpt}, Alpaca \cite{alpaca}, and Vicuna \cite{vicuna2023, zheng2023judging}.
ChatGPT was developed by OpenAI and is one of the state-of-the-art LLMs demonstrating robust abilities for generating human-like text. Our experiments are conducted using the \textit{gpt-3.5-turbo} API.
Alpaca is an open-source LLM with 7B parameters that achieves a good balance between performance and efficiency. It was fine-tuned from the LLaMA-7B model \cite{touvron2023llama} using 52K instructions gathered via a ``self-instruct'' methodology \cite{wang2022self}. 
Vicuna is another open-source LLM trained by fine-tuning the LLaMA2 model on about 125,000 user-shared conversations with ChatGPT from ShareGPT. We use Vicuna v1.5 with 7B parameters. During experiments, we set the temperature of these LLMs to 0 when answering questions and to default when generating stories and rules.

\section{Commonsense Retrieval from LLMs as Stories and Rules}

In this section, we answer the first question: \textit{Which expression, story or rule, is more effective for retrieving commonsense from the memory of LLMs?} First, we compare the confidence in generating stories and commonsense rules. Then, we employ an automatic evaluation to assess the accuracy of commonsense within the generated stories and rules.

\subsection{Confidence of Commonsense Generation} \label{3.1}

To assess the confidence in generating stories and rules using LLMs, we ask LLMs to write corresponding stories of past experiences and commonsense rules given questions from commonsense QA datasets as input (as shown in Figure \ref{f0}).
Specifically, we randomly select 100 questions from each dataset and instruct Alpaca and Vicuna models to generate 5 stories and 5 rules using specific prompts (shown in Table \ref{t1} in Appendix \ref{prompts}).
We use perplexity to indicate the generation confidence of LLMs, which has been a longstanding confidence measure for language models \cite{10.1162/tacl_a_00407}. However, there is a notable difference in word usage between stories and rules. Stories tend to incorporate less common words such as people's names and specific scenes, while rules typically consist of more general and common words. To account for this variation in word frequencies, we subtract the text perplexity with the perplexity of randomly shuffled word lists, which is a common practice in psychological linguistic studies to account for the word frequency effects \cite{humphries2006syntactic, pallier2011cortical, zaccarella2017reviewing, labache2019sentence, Zhang2023ASW}. The confidence is measured by the ``Perplexity Reduction (PR)'':
\begin{equation}
\setlength{\abovedisplayskip}{0.1cm}
\setlength{\belowdisplayskip}{0.1cm}
\label{equation1}
 {\rm PR}(t)={\rm PPL}({\rm shuffle}(t))-{\rm PPL}(t)
\end{equation}
Here, ${\rm PPL}(\cdot)$ denotes the perplexity calculation function, and ${\rm shuffle}(t)$ refers to the shuffling of the text $t$ by words. A higher PR indicates that the LLM is more confident with the text.

\begin{figure}[!t]
  \centering
  \includegraphics[width=\columnwidth]{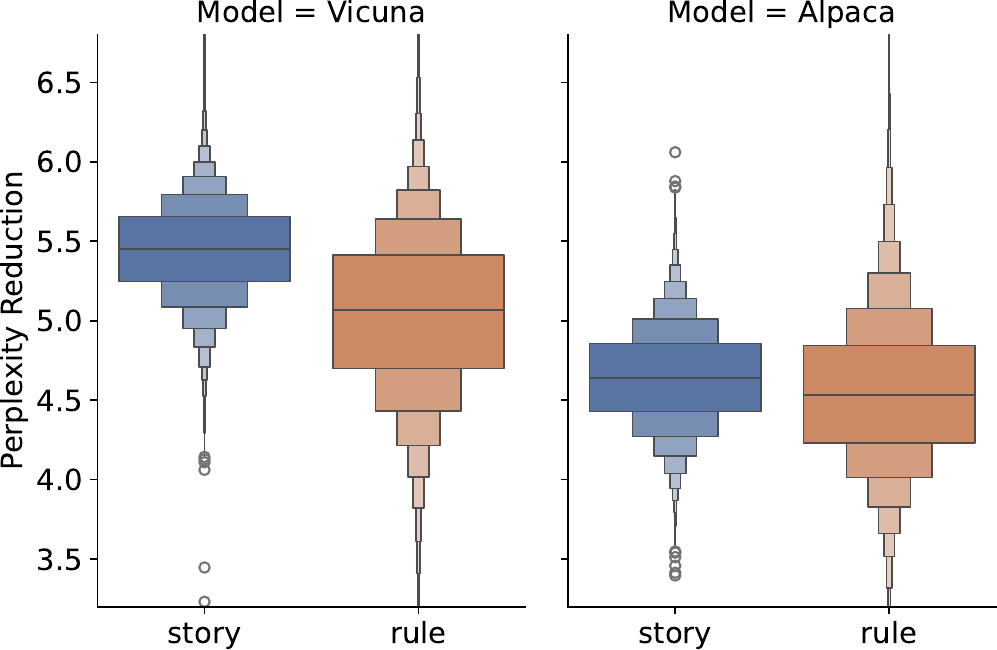}
  \caption{Comparison of perplexity reduction between generating stories and rules. Sample size $N=14,000$ for each setting.}
  \label{f1}
\end{figure}

\paragraph{Finding 1.}
\textbf{When retrieving commonsense from LLMs, stories can result in more confident commonsense generation than rules}.
The results in Figure \ref{f1} show that stories have significantly higher PR than rules for both Vicuna and Alpaca models ($p \ll 0.001$). This effect is more obvious for Vicuna, and we believe this is because Vicuna has better storytelling and instruction-following abilities than Alpaca. 
These observations align with the reporting bias, where commonsense rules are less prevalent in the training text of LLMs, resulting in lower confidence in generating them. Conversely, human language is more likely to convey commonsense as stories, so LLMs develop an ability to generate stories with more confidence.

\subsection{Accuracy of Commonsense Generation} \label{commonsense_acc}

\begin{table}[!t]
\setlength{\belowcaptionskip}{-0.3cm}
\centering
\small
\begin{tabular}{c|ccc}
\hline
Setting & ChatGPT & Vicuna & Alpaca \\
\hline
Story & \textbf{99.42\%} & \textbf{98.82\%} & \textbf{95.39\%} \\
Rule & 98.56\% & 96.21\% & 93.25\% \\
\hline
\end{tabular}
\caption{\label{t2}
Commonsense accuracy of stories and rules.
}
\end{table}

To assess which expression incorporates more accurate commonsense knowledge, we ask ChatGPT to determine if each story or rule aligns with commonsense, responding with either ``yes'' or ``no'' (prompt shown in Table \ref{t1eval} in Appendix \ref{prompts}). We use the same 100 questions for each dataset in Section \ref{3.1}, randomly selecting one story and one rule for evaluation, which results in a total of 2,800 stories and 2,800 rules for each model. It is important to note that these annotations are used solely for the purpose of comparing stories and rules regarding commonsense accuracy under the same conditions, both annotated by ChatGPT, rather than implying absolute accuracy. Further human evaluation results for a small set of stories and rules are shown in Table \ref{t-manual} in Appendix \ref{manual}.

\paragraph{Finding 2.} \textbf{LLMs generate more accurate stories than rules in terms of commonsense.}
Table \ref{t2} shows that the commonsense accuracy of stories is higher than that of rules for all three models ($p<0.0005$). This verifies that the story is a more accurate commonsense expression than the explicit commonsense rule for retrieving commonsense knowledge from LLMs, highlighting the potential of stories as valuable commonsense sources.

\section{Leveraging Commonsense in Stories and Rules for Problem Solving}

This section answers the second question: \textit{Which expression, story or rule, is more suitable for LLMs to leverage commonsense in solving problems?} We compare the confidence of reasoning with stories or rules as contexts. Then, we assess the performance of commonsense QA by employing either stories or rules as contexts and perform detailed analyses.

\subsection{Confidence in Commonsense Reasoning}

To evaluate the confidence of reasoning and generating the correct answer in commonsense QA given stories and rules as contexts, we compare the perplexity reduction of sequence ``context, question (and options if applicable), correct answer'' with context as either stories or rules. Specifically, we employ the same set of 100 questions, the stories, and the rules of each dataset in Section \ref{3.1}. The perplexity reduction is calculated similarly:
\begin{equation}
\begin{aligned}
\setlength{\abovedisplayskip}{0.1cm}
\setlength{\belowdisplayskip}{0.1cm}
\label{equation2}
 &{\rm PR}([c,q,a]) \\
 &= {\rm PPL}([{\rm shuffle}(c),q,a]) - {\rm PPL}([c,q,a])
\end{aligned}
\end{equation}
where $c$, $q$, and $a$ are the context, question, and correct answer, respectively.

\begin{figure}[!t]
\setlength{\belowcaptionskip}{-0.3cm}
  \centering
  \includegraphics[width=\columnwidth]{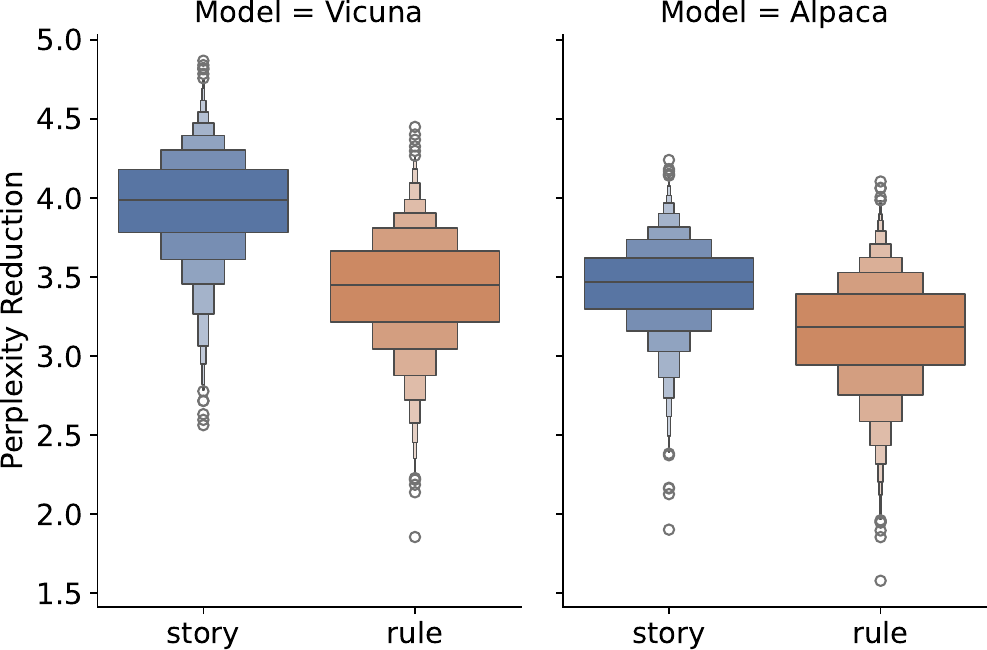}
  \caption{Comparison of perplexity reduction in generating the correct answer with stories or rules as context. Sample size $N=14,000$ for each setting.}
  \label{f2}
\end{figure}

\begin{table*}[!t]
\setlength{\belowcaptionskip}{-0.3cm}
\renewcommand{\arraystretch}{0.91}
\centering
\resizebox{0.99\linewidth}{!}{
\small
\begin{tabular}{l||cccc||cccc||cccc}
\hline
\multirow{2}{*}{Datasets}                 & \multicolumn{4}{c||}{ChatGPT (gpt-3.5-turbo)}                                                                           & \multicolumn{4}{c||}{\textbf{Vicuna}}                                                                                   & \multicolumn{4}{c}{Alpaca}                                                                                            \\
                                          & Base                     & Story                     & Rule                                & Both                      & Base                     & Story                     & Rule                                & Both                      & Base                     & Story                     & Rule                                & Both                     \\ \hline
$^{\dagger}$HellaSWAG-act.net             & \textbf{67.57}           & \underline{60.96}         & \multicolumn{1}{l|}{60.86}          & 62.83                     & 45.75                    & \textbf{48.07}            & \multicolumn{1}{l|}{42.29}          & \textbf{48.40}            & 29.29                    & \textbf{32.23}            & \multicolumn{1}{l|}{29.54}          & \textbf{33.60}           \\
$^{\dagger}$SWAG                          & \textbf{69.50}           & \underline{61.58}         & \multicolumn{1}{l|}{61.22}          & 62.30                     & 48.00                    & \textbf{48.28}            & \multicolumn{1}{l|}{43.92}          & 46.19                     & 30.44                    & \textbf{35.54}            & \multicolumn{1}{l|}{32.42}          & \textbf{36.47}           \\
$^{\dagger}\alpha$NLI                     & 76.21                    & \textbf{77.64}            & \multicolumn{1}{l|}{75.94}          & \textbf{79.28}            & 59.57                    & \textbf{64.51}            & \multicolumn{1}{l|}{61.02}          & \textbf{64.67}            & 59.67                    & \textbf{61.44}            & \multicolumn{1}{l|}{58.89}          & \textbf{62.37}           \\
$^{\dagger}$SCT                           & \textbf{96.48}           & 95.94                     & \multicolumn{1}{l|}{96.26}          & \textbf{97.35}            & 79.82                    & \textbf{82.24}            & \multicolumn{1}{l|}{78.93}          & \textbf{83.60}            & \textbf{86.50}           & \underline{84.15}         & \multicolumn{1}{l|}{83.31}          & 83.57                    \\
QuaRel                                    & 70.04                    & 77.82                     & \multicolumn{1}{l|}{\textbf{80.36}} & 80.29                     & 55.60                    & \textbf{60.89}            & \multicolumn{1}{l|}{59.18}          & 60.15                     & 51.82                    & \textbf{57.55}            & \multicolumn{1}{l|}{54.32}          & \textbf{58.63}           \\
PIQA                                      & 84.20                    & 84.07                     & \multicolumn{1}{l|}{\textbf{84.40}} & \textbf{84.90}            & 65.61                    & \textbf{67.03}            & \multicolumn{1}{l|}{65.39}          & \textbf{67.59}            & 54.25                    & \textbf{54.91}            & \multicolumn{1}{l|}{54.58}          & \textbf{57.82}           \\
WinoGrande                                & 63.38                    & \textbf{70.51}            & \multicolumn{1}{l|}{68.33}          & 69.78                     & 57.74                    & \textbf{60.11}            & \multicolumn{1}{l|}{58.48}          & \textbf{60.28}            & 50.08                    & \textbf{52.53}            & \multicolumn{1}{l|}{50.24}          & 51.42                    \\ \hline
$^{\ddagger}$AI2Sci-elem                 & \textbf{92.68}           & 87.80                     & \multicolumn{1}{l|}{\textbf{92.68}} & 90.24                     & 51.22                    & \textbf{62.81}            & \multicolumn{1}{l|}{61.98}          & \textbf{63.03}            & 42.62                    & 46.72                     & \multicolumn{1}{l|}{\textbf{52.85}} & 47.97                    \\
$^{\dagger}$HellaSWAG-wikihow             & \textbf{73.62}           & \underline{69.45}         & \multicolumn{1}{l|}{62.75}          & 68.29                     & 28.89                    & \textbf{31.44}            & \multicolumn{1}{l|}{30.63}          & 28.69                     & 26.43                    & \textbf{25.78}            & \multicolumn{1}{l|}{25.64}          & 25.42                    \\
CommonsenseQA                             & 74.98                    & \textbf{76.00}            & \multicolumn{1}{l|}{75.84}          & \textbf{77.89}            & 35.60                    & \textbf{47.05}            & \multicolumn{1}{l|}{46.34}          & \textbf{49.01}            & 29.01                    & \textbf{33.94}            & \multicolumn{1}{l|}{33.52}          & \textbf{34.70}           \\ \hline
SocialIQA                                 & 70.21                    & 70.02                     & \multicolumn{1}{l|}{\textbf{70.44}} & \textbf{71.47}            & \textbf{46.24}           & 42.64                     & \multicolumn{1}{l|}{43.66}          & 43.73                     & 42.16                    & \textbf{43.46}            & \multicolumn{1}{l|}{42.69}          & \textbf{44.10}           \\
$^{\ddagger}$ARC-challenge               & 82.27                    & 79.93                     & \multicolumn{1}{l|}{\textbf{84.95}} & 82.94                     & 42.62                    & 50.00                     & \multicolumn{1}{l|}{\textbf{51.54}} & 50.87                     & 36.79                    & \textbf{40.13}            & \multicolumn{1}{l|}{36.36}          & \textbf{40.74}           \\
WSC                                       & 73.33                    & \textbf{82.11}            & \multicolumn{1}{l|}{80.70}          & \textbf{83.03}            & 62.46                    & 63.00                     & \multicolumn{1}{l|}{\textbf{64.94}} & \textbf{65.81}            & 55.48                    & \textbf{56.74}            & \multicolumn{1}{l|}{54.61}          & 55.63                    \\
CODAH                                     & \textbf{85.77}           & 81.47                     & \multicolumn{1}{l|}{82.55}          & 83.09                     & 58.38                    & 56.45                     & \multicolumn{1}{l|}{\textbf{58.63}} & \textbf{59.88}            & \textbf{44.32}           & \underline{41.20}         & \multicolumn{1}{l|}{38.69}          & 41.12                    \\
$^{\dagger\dagger}$Com2Sense              & 70.46                    & 66.28                     & \multicolumn{1}{l|}{\textbf{75.45}} & 70.72                     & 52.30                    & 53.20                     & \multicolumn{1}{l|}{\textbf{55.70}} & 54.74                     & 49.68                    & \textbf{50.45}            & \multicolumn{1}{l|}{50.38}          & 50.38                    \\
WinoVenti                                 & 75.41                    & 77.66                     & \multicolumn{1}{l|}{\textbf{79.09}} & \textbf{79.61}            & 58.46                    & 58.79                     & \multicolumn{1}{l|}{\textbf{61.35}} & 60.11                     & 52.49                    & 53.60                     & \multicolumn{1}{l|}{\textbf{54.57}} & \textbf{56.99}           \\ 
CycIC                                     & 64.26                    & 68.59                     & \multicolumn{1}{l|}{\textbf{74.49}} & 70.62                     & 43.03                    & 43.44                     & \multicolumn{1}{l|}{\textbf{46.02}} & 45.40                     & 33.91                    & 35.94                     & \multicolumn{1}{l|}{\textbf{38.68}} & \textbf{39.61}           \\
QuaRTz                                    & 72.40                    & 77.60                     & \multicolumn{1}{l|}{\textbf{82.81}} & 78.33                     & 52.74                    & 58.52                     & \multicolumn{1}{l|}{\textbf{61.20}} & 59.23                     & 57.72                    & 58.36                     & \multicolumn{1}{l|}{\textbf{65.62}} & 64.83                    \\
$^{\dagger\dagger}$CommonsenseQA2.0       & 64.90                    & 63.15                     & \multicolumn{1}{l|}{\textbf{70.09}} & 65.76                     & 50.68                    & 50.12                     & \multicolumn{1}{l|}{\textbf{53.43}} & 52.90                     & 48.44                    & 48.96                     & \multicolumn{1}{l|}{\textbf{49.55}} & 48.70                    \\
NumerSense                                & 73.50                    & 72.50                     & \multicolumn{1}{l|}{\textbf{76.00}} & \textbf{76.50}            & 47.00                    & 44.00                     & \multicolumn{1}{l|}{\textbf{47.50}} & 47.50                     & 28.00                    & 40.00                     & \multicolumn{1}{l|}{\textbf{54.00}} & 51.50                    \\
$^{\ddagger}$SciQ                         & \textbf{93.30}           & \underline{92.08}         & \multicolumn{1}{l|}{91.98}          & 92.69                     & 65.35                    & 68.19                     & \multicolumn{1}{l|}{\textbf{71.71}} & \textbf{71.87}            & 47.90                    & \textbf{57.26}            & \multicolumn{1}{l|}{54.45}          & 56.40                    \\
$^{\ddagger}$OpenBookQA                   & 78.00                    & \textbf{80.00}            & \multicolumn{1}{l|}{78.92}          & \textbf{82.93}            & 34.80                    & 41.28                     & \multicolumn{1}{l|}{\textbf{44.86}} & \textbf{46.62}            & 33.80                    & 35.27                     & \multicolumn{1}{l|}{\textbf{36.49}} & \textbf{37.15}           \\
COPA                                      & \textbf{96.40}           & 94.30                     & \multicolumn{1}{l|}{95.11}          & 95.26                     & 69.74                    & 75.86                     & \multicolumn{1}{l|}{\textbf{79.71}} & \textbf{82.89}            & \textbf{81.20}           & 75.80                     & \multicolumn{1}{l|}{75.80}          & 79.00                    \\
$^{\ddagger}$QASC                         & 75.70                    & \textbf{77.43}            & \multicolumn{1}{l|}{77.14}          & \textbf{79.14}            & 27.09                    & 40.53                     & \multicolumn{1}{l|}{\textbf{45.08}} & 41.64                     & 22.59                    & \textbf{28.57}            & \multicolumn{1}{l|}{26.14}          & 27.68                    \\
$^{\dagger\dagger}$ComVE (Task A)         & 92.26                    & 87.18                     & \multicolumn{1}{l|}{\textbf{94.71}} & 90.94                     & 49.85                    & 48.59                     & \multicolumn{1}{l|}{\textbf{53.67}} & 48.54                     & 52.77                    & \textbf{56.94}            & \multicolumn{1}{l|}{54.09}          & \textbf{58.33}           \\
$^{\ddagger}$ARC-easy                    & 92.46                    & 92.11                     & \multicolumn{1}{l|}{\textbf{93.86}} & 92.81                     & 59.30                    & 63.15                     & \multicolumn{1}{l|}{\textbf{68.28}} & 65.29                     & 51.67                    & 53.68                     & \multicolumn{1}{l|}{\textbf{54.74}} & \textbf{56.59}           \\
PROST                                     & 53.00                    & 49.90                     & \multicolumn{1}{l|}{\textbf{62.90}} & 53.15                     & 31.29                    & 32.40                     & \multicolumn{1}{l|}{\textbf{38.77}} & 38.06                     & 30.10                    & 31.40                     & \multicolumn{1}{l|}{\textbf{33.50}} & 32.20                    \\
$^{\ddagger}$AI2Sci-middle               & 88.80                    & \textbf{92.00}            & \multicolumn{1}{l|}{90.40}          & 92.00                     & 60.80                    & 59.20                     & \multicolumn{1}{l|}{\textbf{67.48}} & 66.13                     & 52.00                    & \textbf{55.28}            & \multicolumn{1}{l|}{46.77}          & 49.59                    \\ \hline
\end{tabular}}
\caption{\label{t3}
Accuracy (\%) in zero-shot commonsense QA under different settings:
Base - without context,
Story - with stories as context,
Rule - with rules as context,
Both - with both stories and rules as context.
Datasets are sorted and grouped by accuracy differences between using stories and using rules in Vicuna, as depicted in Figure \ref{f-acc}.
${\dagger}$ Datasets related to daily events.
${\ddagger}$ Scientific commonsense datasets.
${\dagger\dagger}$ Datasets related to negation.
}
\end{table*}

\paragraph{Finding 3.} \textbf{LLMs are more confident in commonsense reasoning based on stories than on rules}. 
Figure \ref{f2} shows a significantly higher perplexity reduction when generating the correct answers for commonsense questions using stories than using rules as contexts ($p<0.0003$). This discrepancy reflects the exposure bias in commonsense reasoning: explicit rules are seldom used by people to reason and solve commonsense problems, resulting in more sparse examples in the text corpora for training LLMs. This finding affirms that LLMs can more naturally leverage stories of past experiences for reasoning in commonsense QA.

\subsection{Effectiveness in Commonsense QA}

Next, we assess the accuracy of zero-shot commonsense QA using stories or rules, comparing them with a baseline setting without contextual commonsense. For each question, LLMs first generate five stories or rules, which are then concatenated as context for answering questions using the same model (with prompts shown in Table \ref{t1qa} in Appendix \ref{prompts}). The results are presented in Table \ref{t2}, and the accuracy differences between using stories and rules are shown in Figure \ref{f-acc} for Vicuna and Figure \ref{f-acc-sup} in Appendix \ref{acc-change} for the other two models.

\begin{figure}[!t]
\setlength{\belowcaptionskip}{-0.3cm}
  \centering
  \includegraphics[width=\columnwidth]{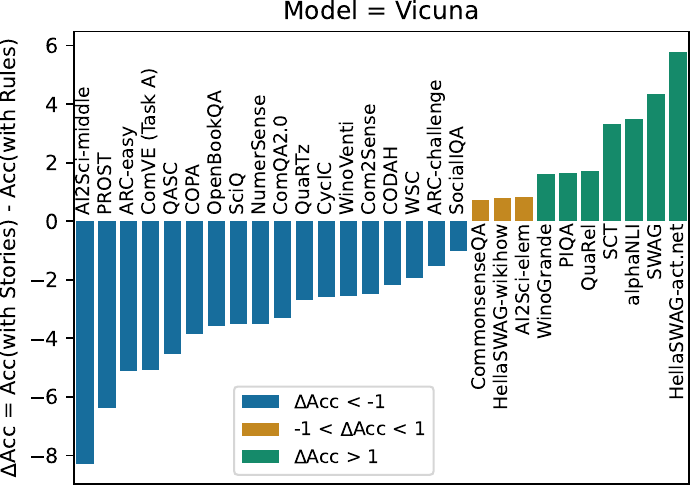}
  \caption{Comparison between the accuracy (\%) with stories and with rules for Vicuna.}
  \label{f-acc}
\end{figure}

\paragraph{Finding 4.}
\textbf{Story is a more effective commonsense expression for answering questions regarding daily events, while the rule is more effective for scientific commonsense QA, which aligns with the reporting bias of commonsense.} 
As shown in Table \ref{t3} and Figure \ref{f-acc}, for datasets like HellaSWAG, SWAG, SCT, and $\alpha$NLI, which involve daily events and stories, leveraging stories as context leads to higher accuracies across all models (only except ChatGPT on the SCT dataset). These datasets involve tasks like selecting correct follow-up behaviors for sequences of events (HellaSWAG and SWAG), choosing appropriate story endings (SCT), or determining events between the beginning and end of a script ($\alpha$NLI). We attribute this to the influence of reporting bias of different types of commonsense in text corpora, which shapes the commonsense ability of LLMs.
Commonsense in our daily life events, such as ``\textit{how to add toothpaste onto a toothbrush}'', is subject to more pronounced reporting bias than other forms of commonsense \cite{shwartz-choi-2020-neural}. This type of commonsense is inherently more implicit and, consequently, more suitable to be expressed as stories.

In contrast, datasets focusing on scientific commonsense at the elementary or middle school level, including OpenBookQA, ARC, QASC, AI2Sci, and SciQ, show better performance when provided with commonsense rules. This is because scientific knowledge is typically structured as rules in textbooks and encyclopedias. The questions and options often include scientific terms, and scientific concepts and descriptions are not commonly presented in the form of stories in the training corpora. 

\paragraph{Finding 5.}
\textbf{Stories and rules can complement each other, further enhancing the QA accuracy of LLMs.}
As shown in Table \ref{t3}, when employing both stories and rules as contextual inputs, LLMs can achieve higher QA accuracy on 10, 12, and 13 datasets compared to using either stories or rules alone. This highlights that the combination of both stories and rules enables LLMs to leverage the unique strengths of each, resulting in a more comprehensive and precise understanding of the presented questions and underlying commonsense. For example, although rules are better for expressing scientific commonsense, stories still play a crucial role in scientific commonsense QA by providing essential contextual information. As shown in Table \ref{t3}, using both stories and rules on the OpenBookQA dataset further improves answer accuracy for all three models.

\subsection{Analyses} \label{analysis}

\subsubsection{Error Analysis} \label{error}

To gain deeper insights into the influence of generated stories, we conduct an error analysis. We focus on questions that are initially answered correctly by the Vicuna model, but the answers change to incorrect when considering the stories as context. We randomly select 10 such questions from each dataset, except for the AI2Sci-elem dataset which has only 8 such questions. Each question comprises five stories, resulting in a total of 1390 stories for error analysis.
We manually classify error cases into 5 primary types: \textbf{Semantic Drifting} (34.0\% – the story drifts away from the question), \textbf{Uncommon or Incorrect Scenarios} (26.6\% – the story does not represent common real-world situations or contains errors), \textbf{Incorrect Answering} (18.6\% – the story is accurate, but the predicted answer is wrong), \textbf{Inconsideration of Options} (16.2\% – the story does not align with any answer options), and \textbf{Inclusion of Wrong Options} (4.6\% – the story emphasizes a wrong answer). The pie chart is shown in Figure \ref{f3} in Appendix \ref{error_chart}.

This analysis highlights two main issues of the stories: commonsense hallucination and semantic drifting. A commonsense hallucination occurs when LLMs are misled by incorrect answer options, generating stories that are uncommon or against commonsense, leading to the uncommon or incorrect scenarios error. Semantic drifting refers to LLMs generating stories whose topics deviate from the question, making them unhelpful in answering the question. These two error types are the primary reasons for incorrect model answers, accounting for over 60\% of total errors. Besides, 18.6\% of the stories are correct and relevant, yet the model fails to effectively use them to answer questions correctly, suggesting room for further improvement in LLMs' ability to leverage contextual information.

\begin{figure}[!t]
\setlength{\belowcaptionskip}{-0.3cm}
  \centering
  \includegraphics[width=\columnwidth]{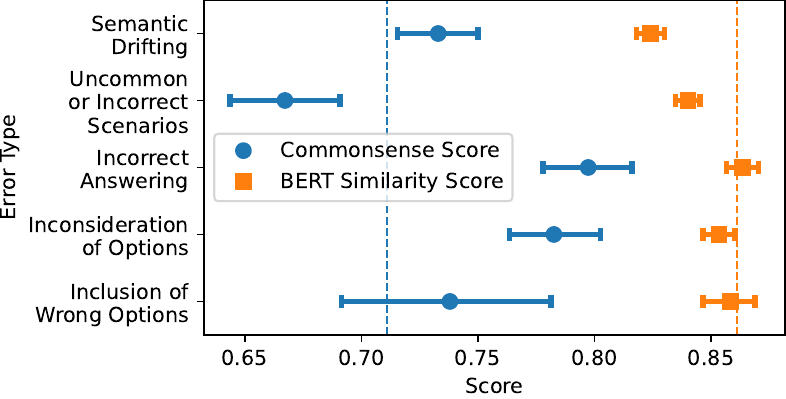}
  \caption{The average scores of stories generated by Vicuna of different error types. The dashed lines are the overall average scores among all questions. Error bars indicate 95\% confidence intervals.}
  \label{f4}
\end{figure}

To quantitatively analyze the generated stories, we employed two scoring methods: commonsense scores and BERT similarity scores. 
For the commonsense score, we use the Vera model \cite{liu-etal-2023-vera}, a T5-based model trained on extensive commonsense statements from knowledge bases. This model outputs a score of the correctness for a given text according to commonsense.
For the BERT similarity score, we calculate the cosine similarity between semantic representations of stories and questions using the BERT-large model \cite{devlin-etal-2019-bert}. The two scores range between 0 and 1. \looseness=-1

There is a correlation between error types and the corresponding scores. Figure \ref{f4} shows that stories with semantic drifting have significantly lower semantic similarities with the questions than the overall average and other error types. Moreover, stories describing uncommon or incorrect scenarios show both lower similarity and commonsense scores in contrast to the overall average and other error types except for semantic drifting. These findings further support the commonsense hallucination and semantic drifting issues.

\subsubsection{Influence of Story on Answer Accuracy}

Further analysis shows notable correlations between answer accuracy and the two scores, the commonsense score and BERT similarity score, at the dataset level (shown in Figure \ref{f5} and \ref{f6} in Appendix \ref{correlation}). Specifically, answer accuracy demonstrates a robust correlation with the commonsense score (Pearson coefficient 0.612, $p<0.001$). In comparison, the correlation with the BERT similarity score is weaker but still positive (Pearson coefficient 0.226, $p=0.003$).
This is because a story involving commonsense hallucination may mislead the model by providing incorrect information, leading to wrong answers, while a story deviating from the question merely offers no relevant information, resulting in a weaker influence. 
These observations underscore the substantial influence of semantic drifting and commonsense hallucination issues on commonsense QA based on stories.

\subsubsection{Analysis of Datasets Related to Negation}

An interesting phenomenon of leveraging commonsense in stories is that LLMs are still not good at handling negations in commonsense. On datasets that involve negation, including CommonsenseQA 2.0 and Com2Sense which require models to assess statement correctness, and ComVE (Task A) which requires identifying statements contradicting commonsense, using rules demonstrates higher accuracy compared to using stories (Table \ref{t3}). Specifically, the Vicuna model tends to more frequently give incorrect ``yes'' responses to questions with a correct answer of ``no'' (69.1\% of error cases in CommonsenseQA 2.0 and 90.2\% in Com2Sense) than the opposite cases. Furthermore, for error cases where the correct answer is ``no'', the model generates stories with more commonsense errors (the average commonsense scores of stories for incorrectly answered ``no'' questions are significantly lower than the opposite with $p<0.003$). This disparity indicates a challenge for LLMs in handling negations within commonsense \cite{chen2023say}. Training corpora include few negative commonsense examples like ``\textit{a stapler is not used for sewing}'', resulting in LLMs generating more hallucinations misled by the given incorrect statements.

\section{Iterative Self-Supervised Fine-Tuning}

After identifying the two issues in generating stories of past experiences, this section presents our iterative self-SFT approach to address these issues. 

\subsection{Method}

Our self-SFT method contains three steps in each iteration: generating, filtering, and training.

In the generating step, we use LLMs to generate stories for questions in the training set. Specifically, we generate five stories for each question.
In the filtering step, we first select the generated stories based on changes in the responses to the questions, i.e., a story is considered helpful if it can rectify an initially incorrect response as correct given it in the context. 
These helpful stories are then scored using a scoring method. The design of the scoring method is crucial for mitigating the commonsense hallucination and semantic drifting issues. We employ the two scores in Section \ref{analysis}: the commonsense score by the Vera model for commonsense correctness and the BERT-based semantic similarity for relevance. The total score is the sum of the two scores.
We retain the $K\%$ top-scored stories for the subsequent training step.
In the training step, we fine-tune an LLM using the filtered stories as output, with inputs following the prompt in Section \ref{3.1}. The fine-tuned model is used for the generating step in the next iteration.

\subsection{Experiments}

\begin{table}[!t]
\setlength{\belowcaptionskip}{-0.1cm}
\centering
\resizebox{\linewidth}{!}{
\small
\begin{tabular}{l|c|ccc}
\hline
Datasets                   & Without SFT    & Iter-1         & Iter-2         & Iter-3         \\ \hline
Seen                       & 51.96          & 52.32          & \textbf{53.26} & 52.12          \\
Unseen                     & 55.30          & \textbf{55.57} & 55.16          & 55.08          \\ \hline
\end{tabular}
}
\caption{\label{t4}
Average accuracy (\%) of commonsense QA by Vicuna with and without iterative self-SFT. Accuracy for each dataset is shown in Table \ref{t7} in Appendix \ref{ablation}.
}
\end{table}

\begin{figure}[!t]
\setlength{\belowcaptionskip}{-0.3cm}
  \centering
  \includegraphics[width=0.90\columnwidth]{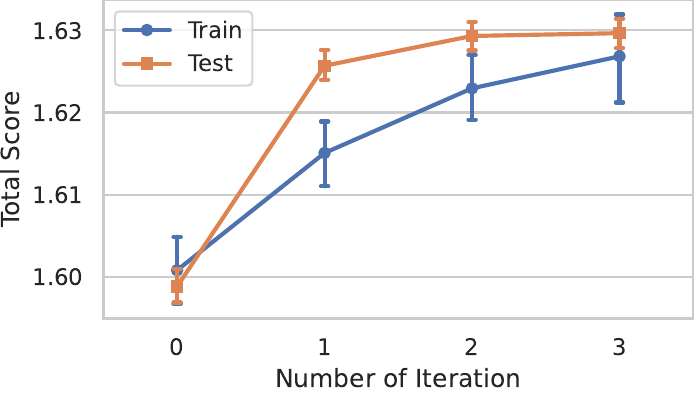}
  \caption{Changes of total score alongside the iteration of self-SFT. Error bars are 95\% confidence intervals.}
  \label{f7}
\end{figure}

Following \citet{liu2023crystal} and our analyses, we train our model on 8 datasets: OpenBookQA, AI2Sci (elementary and middle school set), WinoGrande, HellaSWAG (ActivitiNet and WikiHow set), SWAG, and $\alpha$NLI. The other datasets are used for unseen evaluation. We fine-tune the Vicuna model using LoRA tuning \cite{hu2021lora}, with hyper-parameters in Appendix \ref{parameter}. 

\textbf{The commonsense ability of LLMs can be further self-improved via iterative self-supervised fine-tuning}. Table \ref{t4} shows that on the datasets used for fine-tuning (seen), the average accuracy at all three iterations outperforms the original Vicuna model without SFT. The most significant improvement is at iteration 2, rising from 51.96\% to 53.26\%. This verifies the effectiveness of self-SFT in enhancing the quality of generated stories. Furthermore, self-SFT demonstrates improvements on unseen datasets at iteration 1, suggesting the method's ability to generalize to other commonsense QA datasets not seen during training. Performance decreases at later iterations can be attributed to over-fitting on seen datasets. Ablation and hyper-parameter studies are shown in Appendix \ref{ablation}. 

\textbf{Our approach is effective in addressing the issues and improving the quality of generated stories}. To further assess the effect of our method in mitigating commonsense hallucination and semantic drifting issues during story generation, we analyze the score variations across training iterations. Figure \ref{f7} shows that, along with the training iterations, the total score (sum of commonsense score and BERT similarity) consistently improves for both the training and testing phases. 

\section{Conclusion}

This paper systematically compares stories and rules as commonsense expressions to retrieve and leverage commonsense knowledge in LLMs. Experimental results show that the story is a better expression for retrieving commonsense from LLMs. LLMs generate stories with more confidence and higher commonsense accuracy than rules. Moreover, the story is a more effective commonsense expression for answering questions regarding daily events, while the rule is more effective for scientific commonsense QA. This phenomenon aligns with the reporting bias of commonsense. Stories and rules can complement each other to further enhance answer accuracy. We provide further insights through in-depth analyses, highlighting two challenges in generated stories: commonsense hallucination and semantic drifting. We show that the correctness and relevance of commonsense stories can be improved via iterative self-supervised fine-tuning, underscoring the potential for self-improvement in the commonsense ability of LLMs.

This paper suggests a new perspective, going beyond the common practice of expressing commonsense as rules. Our results and findings emphasize the importance of using the appropriate language to express, retrieve, and leverage commonsense for LLMs to further exploit their potential. The full extent of LLMs in handling commonsense is not yet fully realized, calling for future research to refine and improve the commonsense abilities of LLMs.

\section*{Limitations}

This study specifically investigates several popular LLMs, including ChatGPT, Vicuna, and Alpaca, while excluding the exploration of other LLMs such as GPT-4, Mistral \cite{jiang2023mistral}, and Google's Bard \cite{thoppilan2022lamda}. The selection of models is based on considerations of popularity, availability, and cost. Future research could provide valuable insights by examining whether similar findings hold for these models and conducting performance comparisons with the models included in this study.

The assessment of commonsense accuracy in Section \ref{commonsense_acc} relies on automatic labeling by ChatGPT. Therefore, the accuracy presented in Table \ref{t2} serves solely for comparing stories and rules, and should not be regarded as absolute accuracy. To ensure the quality of this automatic evaluation, we conduct a manual evaluation of a small set of stories and rules, as shown in Table \ref{t-manual} in Appendix \ref{manual}. However, due to its labor-intensive nature, manually labeling all generated stories and rules is impractical for us. Future studies should incorporate more extensive human evaluations to provide a more comprehensive and nuanced understanding of the commonsense knowledge generated by LLMs.

This paper employs two scores to assess the quality of generated stories and to filter training data. However, these two scores—the commonsense score and BERT similarity score—may inherently exhibit biases as they rely on model-based scoring methods. For instance, the Vera model for commonsense score is fine-tuned from the T5 model using commonsense statements synthesized from commonsense knowledge bases, potentially leading the score to favor statements aligned with these knowledge bases. We anticipate detailed analyses of potential biases in these models and scores.

Lastly, in the manual error analysis process, we discover that some commonsense QA datasets are not actually asking about commonsense, despite being recognized as commonsense questions. Some datasets are automatically constructed based on knowledge graphs, probably leading to unreasonable questions or insufficient information to answer the question. Other manually constructed datasets may face the challenge that different annotators may have entirely different understandings of what commonsense is. We follow the common practices in commonsense studies and use these datasets in our commonsense QA experiments. Further investigation into the existing commonsense QA datasets is a task for future studies.

\section*{Acknowledgements}
We sincerely thank all anonymous reviewers for their insightful comments and valuable suggestions. This research work is supported by the Strategic Priority Research Program of Chinese Academy of Sciences, Grant No. XDA27020200, and the National Natural Science Foundation of China under Grants No. 62122077, 62106251, and 62306303.

\bibliography{commonsense}

\begin{thebibliography}{107}
\expandafter\ifx\csname natexlab\endcsname\relax\def\natexlab#1{#1}\fi

\bibitem[{AllenAI(2017)}]{AI2Sci}
AllenAI. 2017.
\newblock Ai2 science questions v2.1 (october 2017).
\newblock \url{http://data.allenai.org/ai2-science-questions/}.

\bibitem[{Aroca-Ouellette et~al.(2021)Aroca-Ouellette, Paik, Roncone, and
  Kann}]{aroca-ouellette-etal-2021-prost}
St{\'e}phane Aroca-Ouellette, Cory Paik, Alessandro Roncone, and Katharina
  Kann. 2021.
\newblock \href {https://doi.org/10.18653/v1/2021.findings-acl.404} {{PROST}:
  {P}hysical reasoning about objects through space and time}.
\newblock In \emph{Findings of the Association for Computational Linguistics:
  ACL-IJCNLP 2021}, pages 4597--4608, Online. Association for Computational
  Linguistics.

\bibitem[{Bhagavatula et~al.(2020)Bhagavatula, Bras, Malaviya, Sakaguchi,
  Holtzman, Rashkin, Downey, Yih, and Choi}]{bhagavatula2019abductive}
Chandra Bhagavatula, Ronan~Le Bras, Chaitanya Malaviya, Keisuke Sakaguchi, Ari
  Holtzman, Hannah Rashkin, Doug Downey, Wen{-}tau Yih, and Yejin Choi. 2020.
\newblock \href {https://openreview.net/forum?id=Byg1v1HKDB} {Abductive
  commonsense reasoning}.
\newblock In \emph{8th International Conference on Learning Representations,
  {ICLR} 2020, Addis Ababa, Ethiopia, April 26-30, 2020}. OpenReview.net.

\bibitem[{Bhandari and Brennan(2023)}]{bhandari2023trustworthiness}
Prabin Bhandari and Hannah~Marie Brennan. 2023.
\newblock \href {https://arxiv.org/abs/2308.00073} {Trustworthiness of children
  stories generated by large language models}.
\newblock \emph{ArXiv preprint}, abs/2308.00073.

\bibitem[{Bian et~al.(2021)Bian, Han, Chen, and Sun}]{bian2021benchmarking}
Ning Bian, Xianpei Han, Bo~Chen, and Le~Sun. 2021.
\newblock \href {https://ojs.aaai.org/index.php/AAAI/article/view/17490}
  {Benchmarking knowledge-enhanced commonsense question answering via
  knowledge-to-text transformation}.
\newblock In \emph{Thirty-Fifth {AAAI} Conference on Artificial Intelligence,
  {AAAI} 2021, Thirty-Third Conference on Innovative Applications of Artificial
  Intelligence, {IAAI} 2021, The Eleventh Symposium on Educational Advances in
  Artificial Intelligence, {EAAI} 2021, Virtual Event, February 2-9, 2021},
  pages 12574--12582. {AAAI} Press.

\bibitem[{Bian et~al.(2023)Bian, Han, Sun, Lin, Lu, and He}]{bian2023chatgpt}
Ning Bian, Xianpei Han, Le~Sun, Hongyu Lin, Yaojie Lu, and Ben He. 2023.
\newblock \href {https://arxiv.org/abs/2303.16421} {Chatgpt is a knowledgeable
  but inexperienced solver: An investigation of commonsense problem in large
  language models}.
\newblock \emph{ArXiv preprint}, abs/2303.16421.

\bibitem[{Bisk et~al.(2020)Bisk, Zellers, LeBras, Gao, and Choi}]{bisk2020piqa}
Yonatan Bisk, Rowan Zellers, Ronan LeBras, Jianfeng Gao, and Yejin Choi. 2020.
\newblock \href {https://aaai.org/ojs/index.php/AAAI/article/view/6239}
  {{PIQA:} reasoning about physical commonsense in natural language}.
\newblock In \emph{The Thirty-Fourth {AAAI} Conference on Artificial
  Intelligence, {AAAI} 2020, The Thirty-Second Innovative Applications of
  Artificial Intelligence Conference, {IAAI} 2020, The Tenth {AAAI} Symposium
  on Educational Advances in Artificial Intelligence, {EAAI} 2020, New York,
  NY, USA, February 7-12, 2020}, pages 7432--7439. {AAAI} Press.

\bibitem[{Bosselut et~al.(2021)Bosselut, Bras, and Choi}]{bosselut2021dynamic}
Antoine Bosselut, Ronan~Le Bras, and Yejin Choi. 2021.
\newblock \href {https://ojs.aaai.org/index.php/AAAI/article/view/16625}
  {Dynamic neuro-symbolic knowledge graph construction for zero-shot
  commonsense question answering}.
\newblock In \emph{Thirty-Fifth {AAAI} Conference on Artificial Intelligence,
  {AAAI} 2021, Thirty-Third Conference on Innovative Applications of Artificial
  Intelligence, {IAAI} 2021, The Eleventh Symposium on Educational Advances in
  Artificial Intelligence, {EAAI} 2021, Virtual Event, February 2-9, 2021},
  pages 4923--4931. {AAAI} Press.

\bibitem[{Brachman and Levesque(2023)}]{brachman2023machines}
R.J. Brachman and H.J. Levesque. 2023.
\newblock \href {https://books.google.co.jp/books?id=igGoEAAAQBAJ}
  {\emph{Machines like Us: Toward AI with Common Sense}}.
\newblock MIT Press.

\bibitem[{Brachman and Levesque(2022)}]{brachman2022toward}
Ronald~J. Brachman and Hector~J. Levesque. 2022.
\newblock \href {https://ojs.aaai.org/index.php/AAAI/article/view/21485}
  {Toward a new science of common sense}.
\newblock In \emph{Thirty-Sixth {AAAI} Conference on Artificial Intelligence,
  {AAAI} 2022, Thirty-Fourth Conference on Innovative Applications of
  Artificial Intelligence, {IAAI} 2022, The Twelveth Symposium on Educational
  Advances in Artificial Intelligence, {EAAI} 2022 Virtual Event, February 22 -
  March 1, 2022}, pages 12245--12249. {AAAI} Press.

\bibitem[{Brown et~al.(2020)Brown, Mann, Ryder, Subbiah, Kaplan, Dhariwal,
  Neelakantan, Shyam, Sastry, Askell, Agarwal, Herbert{-}Voss, Krueger,
  Henighan, Child, Ramesh, Ziegler, Wu, Winter, Hesse, Chen, Sigler, Litwin,
  Gray, Chess, Clark, Berner, McCandlish, Radford, Sutskever, and
  Amodei}]{brown2020language}
Tom~B. Brown, Benjamin Mann, Nick Ryder, Melanie Subbiah, Jared Kaplan,
  Prafulla Dhariwal, Arvind Neelakantan, Pranav Shyam, Girish Sastry, Amanda
  Askell, Sandhini Agarwal, Ariel Herbert{-}Voss, Gretchen Krueger, Tom
  Henighan, Rewon Child, Aditya Ramesh, Daniel~M. Ziegler, Jeffrey Wu, Clemens
  Winter, Christopher Hesse, Mark Chen, Eric Sigler, Mateusz Litwin, Scott
  Gray, Benjamin Chess, Jack Clark, Christopher Berner, Sam McCandlish, Alec
  Radford, Ilya Sutskever, and Dario Amodei. 2020.
\newblock \href
  {https://proceedings.neurips.cc/paper/2020/hash/1457c0d6bfcb4967418bfb8ac142f64a-Abstract.html}
  {Language models are few-shot learners}.
\newblock In \emph{Advances in Neural Information Processing Systems 33: Annual
  Conference on Neural Information Processing Systems 2020, NeurIPS 2020,
  December 6-12, 2020, virtual}.

\bibitem[{Cassirer et~al.(1946)Cassirer, Cassirer, Langer, Hansen-Love, and
  Mari{\'c}}]{cassirer1946language}
E.~Cassirer, E.A. Cassirer, S.K.K. Langer, O.~Hansen-Love, and S.~Mari{\'c}.
  1946.
\newblock \emph{Language and Myth}.
\newblock Dover Books on Literature, Philosophy, History, Religion. Dover
  Publications.

\bibitem[{Chang et~al.(2020)Chang, Liu, Gopalakrishnan, Hedayatnia, Zhou, and
  Hakkani-Tur}]{chang-etal-2020-incorporating}
Ting-Yun Chang, Yang Liu, Karthik Gopalakrishnan, Behnam Hedayatnia, Pei Zhou,
  and Dilek Hakkani-Tur. 2020.
\newblock \href {https://doi.org/10.18653/v1/2020.deelio-1.9} {Incorporating
  commonsense knowledge graph in pretrained models for social commonsense
  tasks}.
\newblock In \emph{Proceedings of Deep Learning Inside Out (DeeLIO): The First
  Workshop on Knowledge Extraction and Integration for Deep Learning
  Architectures}, pages 74--79, Online. Association for Computational
  Linguistics.

\bibitem[{Chen et~al.(2023{\natexlab{a}})Chen, Shi, Fu, Cheng, Li, and
  Xiao}]{chen2023say}
Jiangjie Chen, Wei Shi, Ziquan Fu, Sijie Cheng, Lei Li, and Yanghua Xiao.
  2023{\natexlab{a}}.
\newblock \href {https://arxiv.org/abs/2305.05976} {Say what you mean! large
  language models speak too positively about negative commonsense knowledge}.
\newblock \emph{ArXiv preprint}, abs/2305.05976.

\bibitem[{Chen et~al.(2019)Chen, D{'}Arcy, Liu, Fernandez, and
  Downey}]{chen-etal-2019-codah}
Michael Chen, Mike D{'}Arcy, Alisa Liu, Jared Fernandez, and Doug Downey. 2019.
\newblock \href {https://doi.org/10.18653/v1/W19-2008} {{CODAH}: An
  adversarially-authored question answering dataset for common sense}.
\newblock In \emph{Proceedings of the 3rd Workshop on Evaluating Vector Space
  Representations for {NLP}}, pages 63--69, Minneapolis, USA. Association for
  Computational Linguistics.

\bibitem[{Chen et~al.(2023{\natexlab{b}})Chen, Xu, Yan, Zhang, Huang, Si, and
  Zhang}]{chen-etal-2023-distinguish}
Qianglong Chen, Guohai Xu, Ming Yan, Ji~Zhang, Fei Huang, Luo Si, and Yin
  Zhang. 2023{\natexlab{b}}.
\newblock \href {https://doi.org/10.18653/v1/2023.findings-acl.835}
  {Distinguish before answer: Generating contrastive explanation as knowledge
  for commonsense question answering}.
\newblock In \emph{Findings of the Association for Computational Linguistics:
  ACL 2023}, pages 13207--13224, Toronto, Canada. Association for Computational
  Linguistics.

\bibitem[{Chiang et~al.(2023)Chiang, Li, Lin, Sheng, Wu, Zhang, Zheng, Zhuang,
  Zhuang, Gonzalez, Stoica, and Xing}]{vicuna2023}
Wei-Lin Chiang, Zhuohan Li, Zi~Lin, Ying Sheng, Zhanghao Wu, Hao Zhang, Lianmin
  Zheng, Siyuan Zhuang, Yonghao Zhuang, Joseph~E. Gonzalez, Ion Stoica, and
  Eric~P. Xing. 2023.
\newblock \href {https://lmsys.org/blog/2023-03-30-vicuna/} {Vicuna: An
  open-source chatbot impressing gpt-4 with 90\%* chatgpt quality}.

\bibitem[{Clark et~al.(2018)Clark, Cowhey, Etzioni, Khot, Sabharwal, Schoenick,
  and Tafjord}]{clark2018think}
Peter Clark, Isaac Cowhey, Oren Etzioni, Tushar Khot, Ashish Sabharwal, Carissa
  Schoenick, and Oyvind Tafjord. 2018.
\newblock \href {https://arxiv.org/abs/1803.05457} {Think you have solved
  question answering? try arc, the ai2 reasoning challenge}.
\newblock \emph{ArXiv preprint}, abs/1803.05457.

\bibitem[{Devlin et~al.(2019)Devlin, Chang, Lee, and
  Toutanova}]{devlin-etal-2019-bert}
Jacob Devlin, Ming-Wei Chang, Kenton Lee, and Kristina Toutanova. 2019.
\newblock \href {https://doi.org/10.18653/v1/N19-1423} {{BERT}: Pre-training of
  deep bidirectional transformers for language understanding}.
\newblock In \emph{Proceedings of the 2019 Conference of the North {A}merican
  Chapter of the Association for Computational Linguistics: Human Language
  Technologies, Volume 1 (Long and Short Papers)}, pages 4171--4186,
  Minneapolis, Minnesota. Association for Computational Linguistics.

\bibitem[{Do and Pavlick(2021)}]{do-pavlick-2021-rotten}
Nam Do and Ellie Pavlick. 2021.
\newblock \href {https://doi.org/10.18653/v1/2021.findings-acl.181} {Are rotten
  apples edible? challenging commonsense inference ability with exceptions}.
\newblock In \emph{Findings of the Association for Computational Linguistics:
  ACL-IJCNLP 2021}, pages 2061--2073, Online. Association for Computational
  Linguistics.

\bibitem[{Eldan and Li(2023)}]{eldan2023tinystories}
Ronen Eldan and Yuanzhi Li. 2023.
\newblock \href {https://arxiv.org/abs/2305.07759} {Tinystories: How small can
  language models be and still speak coherent english?}
\newblock \emph{ArXiv preprint}, abs/2305.07759.

\bibitem[{Gordon and Van~Durme(2013)}]{gordon2013reporting}
Jonathan Gordon and Benjamin Van~Durme. 2013.
\newblock Reporting bias and knowledge acquisition.
\newblock In \emph{Proceedings of the 2013 workshop on Automated knowledge base
  construction}, pages 25--30.

\bibitem[{Gu et~al.(2022)Gu, Dalvi, and Clark}]{gu-etal-2022-dream}
Yuling Gu, Bhavana Dalvi, and Peter Clark. 2022.
\newblock \href {https://doi.org/10.18653/v1/2022.naacl-main.82} {{DREAM}:
  Improving situational {QA} by first elaborating the situation}.
\newblock In \emph{Proceedings of the 2022 Conference of the North American
  Chapter of the Association for Computational Linguistics: Human Language
  Technologies}, pages 1115--1127, Seattle, United States. Association for
  Computational Linguistics.

\bibitem[{Hu et~al.(2022)Hu, Shen, Wallis, Allen{-}Zhu, Li, Wang, Wang, and
  Chen}]{hu2021lora}
Edward~J. Hu, Yelong Shen, Phillip Wallis, Zeyuan Allen{-}Zhu, Yuanzhi Li,
  Shean Wang, Lu~Wang, and Weizhu Chen. 2022.
\newblock \href {https://openreview.net/forum?id=nZeVKeeFYf9} {Lora: Low-rank
  adaptation of large language models}.
\newblock In \emph{The Tenth International Conference on Learning
  Representations, {ICLR} 2022, Virtual Event, April 25-29, 2022}.
  OpenReview.net.

\bibitem[{Humphries et~al.(2006)Humphries, Binder, Medler, and
  Liebenthal}]{humphries2006syntactic}
Colin Humphries, Jeffrey~R Binder, David~A Medler, and Einat Liebenthal. 2006.
\newblock Syntactic and semantic modulation of neural activity during auditory
  sentence comprehension.
\newblock \emph{Journal of cognitive neuroscience}, 18(4):665--679.

\bibitem[{Hwang et~al.(2021)Hwang, Bhagavatula, Bras, Da, Sakaguchi, Bosselut,
  and Choi}]{hwang2021comet}
Jena~D. Hwang, Chandra Bhagavatula, Ronan~Le Bras, Jeff Da, Keisuke Sakaguchi,
  Antoine Bosselut, and Yejin Choi. 2021.
\newblock \href {https://ojs.aaai.org/index.php/AAAI/article/view/16792}
  {(comet-) atomic 2020: On symbolic and neural commonsense knowledge graphs}.
\newblock In \emph{Thirty-Fifth {AAAI} Conference on Artificial Intelligence,
  {AAAI} 2021, Thirty-Third Conference on Innovative Applications of Artificial
  Intelligence, {IAAI} 2021, The Eleventh Symposium on Educational Advances in
  Artificial Intelligence, {EAAI} 2021, Virtual Event, February 2-9, 2021},
  pages 6384--6392. {AAAI} Press.

\bibitem[{Ilievski et~al.(2021)Ilievski, Szekely, and Zhang}]{ilievski2021cskg}
Filip Ilievski, Pedro Szekely, and Bin Zhang. 2021.
\newblock Cskg: The commonsense knowledge graph.
\newblock In \emph{The Semantic Web: 18th International Conference, ESWC 2021,
  Virtual Event, June 6--10, 2021, Proceedings 18}, pages 680--696. Springer.

\bibitem[{Ismayilzada et~al.(2023)Ismayilzada, Paul, Montariol, Geva, and
  Bosselut}]{ismayilzada2023crow}
Mete Ismayilzada, Debjit Paul, Syrielle Montariol, Mor Geva, and Antoine
  Bosselut. 2023.
\newblock \href {https://arxiv.org/abs/2310.15239} {Crow: Benchmarking
  commonsense reasoning in real-world tasks}.
\newblock \emph{ArXiv preprint}, abs/2310.15239.

\bibitem[{Jiang et~al.(2023)Jiang, Sablayrolles, Mensch, Bamford, Chaplot,
  Casas, Bressand, Lengyel, Lample, Saulnier et~al.}]{jiang2023mistral}
Albert~Q Jiang, Alexandre Sablayrolles, Arthur Mensch, Chris Bamford,
  Devendra~Singh Chaplot, Diego de~las Casas, Florian Bressand, Gianna Lengyel,
  Guillaume Lample, Lucile Saulnier, et~al. 2023.
\newblock \href {https://arxiv.org/abs/2310.06825} {Mistral 7b}.
\newblock \emph{ArXiv preprint}, abs/2310.06825.

\bibitem[{Jiang et~al.(2021)Jiang, Araki, Ding, and
  Neubig}]{10.1162/tacl_a_00407}
Zhengbao Jiang, Jun Araki, Haibo Ding, and Graham Neubig. 2021.
\newblock \href {https://doi.org/10.1162/tacl_a_00407} {{How Can We Know When
  Language Models Know? On the Calibration of Language Models for Question
  Answering}}.
\newblock \emph{Transactions of the Association for Computational Linguistics},
  9:962--977.

\bibitem[{Jiayang et~al.(2023)Jiayang, Qiu, Chan, Fang, Wang, Chan, Ru, Guo,
  Zhang, Song, Zhang, and Zhang}]{jiayang-etal-2023-storyanalogy}
Cheng Jiayang, Lin Qiu, Tsz Chan, Tianqing Fang, Weiqi Wang, Chunkit Chan,
  Dongyu Ru, Qipeng Guo, Hongming Zhang, Yangqiu Song, Yue Zhang, and Zheng
  Zhang. 2023.
\newblock \href {https://doi.org/10.18653/v1/2023.emnlp-main.706}
  {{S}tory{A}nalogy: Deriving story-level analogies from large language models
  to unlock analogical understanding}.
\newblock In \emph{Proceedings of the 2023 Conference on Empirical Methods in
  Natural Language Processing}, pages 11518--11537, Singapore. Association for
  Computational Linguistics.

\bibitem[{Khot et~al.(2020)Khot, Clark, Guerquin, Jansen, and
  Sabharwal}]{khot2020qasc}
Tushar Khot, Peter Clark, Michal Guerquin, Peter Jansen, and Ashish Sabharwal.
  2020.
\newblock \href {https://aaai.org/ojs/index.php/AAAI/article/view/6319}
  {{QASC:} {A} dataset for question answering via sentence composition}.
\newblock In \emph{The Thirty-Fourth {AAAI} Conference on Artificial
  Intelligence, {AAAI} 2020, The Thirty-Second Innovative Applications of
  Artificial Intelligence Conference, {IAAI} 2020, The Tenth {AAAI} Symposium
  on Educational Advances in Artificial Intelligence, {EAAI} 2020, New York,
  NY, USA, February 7-12, 2020}, pages 8082--8090. {AAAI} Press.

\bibitem[{Klein(2004)}]{gary2004power}
Gary Klein. 2004.
\newblock \href {https://books.google.co.jp/books?id=_rVEardwd6wC} {\emph{The
  Power of Intuition: How to Use Your Gut Feelings to Make Better Decisions at
  Work}}.
\newblock Currency/Doubleday.

\bibitem[{Labache et~al.(2019)Labache, Joliot, Saracco, Jobard, Hesling, Zago,
  Mellet, Petit, Crivello, Mazoyer et~al.}]{labache2019sentence}
Loic Labache, Marc Joliot, J{\'e}r{\^o}me Saracco, Ga{\"e}l Jobard, Isabelle
  Hesling, Laure Zago, Emmanuel Mellet, Laurent Petit, Fabrice Crivello,
  Bernard Mazoyer, et~al. 2019.
\newblock A sentence supramodal areas atlas (sensaas) based on multiple
  task-induced activation mapping and graph analysis of intrinsic connectivity
  in 144 healthy right-handers.
\newblock \emph{Brain Structure and Function}, 224(2):859--882.

\bibitem[{Lal et~al.(2022)Lal, Tandon, Aggarwal, Liu, Chambers, Mooney, and
  Balasubramanian}]{lal2022using}
Yash~Kumar Lal, Niket Tandon, Tanvi Aggarwal, Horace Liu, Nathanael Chambers,
  Raymond Mooney, and Niranjan Balasubramanian. 2022.
\newblock Using commonsense knowledge to answer why-questions.
\newblock In \emph{Proceedings of the 2022 Conference on Empirical Methods in
  Natural Language Processing}, pages 1204--1219.

\bibitem[{Latcinnik and Berant(2020)}]{latcinnik2020explaining}
Veronica Latcinnik and Jonathan Berant. 2020.
\newblock \href {https://arxiv.org/abs/2004.05569} {Explaining question
  answering models through text generation}.
\newblock \emph{ArXiv preprint}, abs/2004.05569.

\bibitem[{Levesque et~al.(2012)Levesque, Davis, and
  Morgenstern}]{levesque2012winograd}
Hector Levesque, Ernest Davis, and Leora Morgenstern. 2012.
\newblock The winograd schema challenge.
\newblock In \emph{Thirteenth international conference on the principles of
  knowledge representation and reasoning}.

\bibitem[{Li et~al.(2022)Li, Kuncoro, Hoffmann, de~Masson~d{'}Autume, Blunsom,
  and Nematzadeh}]{li-etal-2022-systematic}
Xiang~Lorraine Li, Adhiguna Kuncoro, Jordan Hoffmann, Cyprien
  de~Masson~d{'}Autume, Phil Blunsom, and Aida Nematzadeh. 2022.
\newblock \href {https://aclanthology.org/2022.emnlp-main.812} {A systematic
  investigation of commonsense knowledge in large language models}.
\newblock In \emph{Proceedings of the 2022 Conference on Empirical Methods in
  Natural Language Processing}, pages 11838--11855, Abu Dhabi, United Arab
  Emirates. Association for Computational Linguistics.

\bibitem[{Li et~al.(2023)Li, Peng, He, Galley, Gao, and Yan}]{li2023guiding}
Zekun Li, Baolin Peng, Pengcheng He, Michel Galley, Jianfeng Gao, and Xifeng
  Yan. 2023.
\newblock \href {https://arxiv.org/abs/2302.11520} {Guiding large language
  models via directional stimulus prompting}.
\newblock \emph{ArXiv preprint}, abs/2302.11520.

\bibitem[{Lin et~al.(2019)Lin, Chen, Chen, and Ren}]{lin-etal-2019-kagnet}
Bill~Yuchen Lin, Xinyue Chen, Jamin Chen, and Xiang Ren. 2019.
\newblock \href {https://doi.org/10.18653/v1/D19-1282} {{K}ag{N}et:
  Knowledge-aware graph networks for commonsense reasoning}.
\newblock In \emph{Proceedings of the 2019 Conference on Empirical Methods in
  Natural Language Processing and the 9th International Joint Conference on
  Natural Language Processing (EMNLP-IJCNLP)}, pages 2829--2839, Hong Kong,
  China. Association for Computational Linguistics.

\bibitem[{Lin et~al.(2020)Lin, Lee, Khanna, and Ren}]{lin-etal-2020-birds}
Bill~Yuchen Lin, Seyeon Lee, Rahul Khanna, and Xiang Ren. 2020.
\newblock \href {https://doi.org/10.18653/v1/2020.emnlp-main.557} {{B}irds have
  four legs?! {N}umer{S}ense: {P}robing {N}umerical {C}ommonsense {K}nowledge
  of {P}re-{T}rained {L}anguage {M}odels}.
\newblock In \emph{Proceedings of the 2020 Conference on Empirical Methods in
  Natural Language Processing (EMNLP)}, pages 6862--6868, Online. Association
  for Computational Linguistics.

\bibitem[{Liu et~al.(2022{\natexlab{a}})Liu, Hallinan, Lu, He, Welleck,
  Hajishirzi, and Choi}]{liu-etal-2022-rainier}
Jiacheng Liu, Skyler Hallinan, Ximing Lu, Pengfei He, Sean Welleck, Hannaneh
  Hajishirzi, and Yejin Choi. 2022{\natexlab{a}}.
\newblock \href {https://aclanthology.org/2022.emnlp-main.611} {Rainier:
  Reinforced knowledge introspector for commonsense question answering}.
\newblock In \emph{Proceedings of the 2022 Conference on Empirical Methods in
  Natural Language Processing}, pages 8938--8958, Abu Dhabi, United Arab
  Emirates. Association for Computational Linguistics.

\bibitem[{Liu et~al.(2022{\natexlab{b}})Liu, Liu, Lu, Welleck, West, Le~Bras,
  Choi, and Hajishirzi}]{liu2022generated}
Jiacheng Liu, Alisa Liu, Ximing Lu, Sean Welleck, Peter West, Ronan Le~Bras,
  Yejin Choi, and Hannaneh Hajishirzi. 2022{\natexlab{b}}.
\newblock \href {https://doi.org/10.18653/v1/2022.acl-long.225} {Generated
  knowledge prompting for commonsense reasoning}.
\newblock In \emph{Proceedings of the 60th Annual Meeting of the Association
  for Computational Linguistics (Volume 1: Long Papers)}, pages 3154--3169,
  Dublin, Ireland. Association for Computational Linguistics.

\bibitem[{Liu et~al.(2023{\natexlab{a}})Liu, Pasunuru, Hajishirzi, Choi, and
  Celikyilmaz}]{liu2023crystal}
Jiacheng Liu, Ramakanth Pasunuru, Hannaneh Hajishirzi, Yejin Choi, and Asli
  Celikyilmaz. 2023{\natexlab{a}}.
\newblock \href {https://arxiv.org/abs/2310.04921} {Crystal: Introspective
  reasoners reinforced with self-feedback}.
\newblock \emph{ArXiv preprint}, abs/2310.04921.

\bibitem[{Liu et~al.(2023{\natexlab{b}})Liu, Wang, Wang, Smith, Choi, and
  Hajishirzi}]{liu-etal-2023-vera}
Jiacheng Liu, Wenya Wang, Dianzhuo Wang, Noah Smith, Yejin Choi, and Hannaneh
  Hajishirzi. 2023{\natexlab{b}}.
\newblock \href {https://doi.org/10.18653/v1/2023.emnlp-main.81} {Vera: A
  general-purpose plausibility estimation model for commonsense statements}.
\newblock In \emph{Proceedings of the 2023 Conference on Empirical Methods in
  Natural Language Processing}, pages 1264--1287, Singapore. Association for
  Computational Linguistics.

\bibitem[{Liu et~al.(2023{\natexlab{c}})Liu, Li, Zhang, Du, Chen, Hu, Xu, Chen,
  and Wu}]{liu2023mind}
Weize Liu, Guocong Li, Kai Zhang, Bang Du, Qiyuan Chen, Xuming Hu, Hongxia Xu,
  Jintai Chen, and Jian Wu. 2023{\natexlab{c}}.
\newblock \href {https://arxiv.org/abs/2311.09214} {Mind's mirror: Distilling
  self-evaluation capability and comprehensive thinking from large language
  models}.
\newblock \emph{ArXiv preprint}, abs/2311.09214.

\bibitem[{Lv et~al.(2020)Lv, Guo, Xu, Tang, Duan, Gong, Shou, Jiang, Cao, and
  Hu}]{lv2020graph}
Shangwen Lv, Daya Guo, Jingjing Xu, Duyu Tang, Nan Duan, Ming Gong, Linjun
  Shou, Daxin Jiang, Guihong Cao, and Songlin Hu. 2020.
\newblock \href {https://aaai.org/ojs/index.php/AAAI/article/view/6364}
  {Graph-based reasoning over heterogeneous external knowledge for commonsense
  question answering}.
\newblock In \emph{The Thirty-Fourth {AAAI} Conference on Artificial
  Intelligence, {AAAI} 2020, The Thirty-Second Innovative Applications of
  Artificial Intelligence Conference, {IAAI} 2020, The Tenth {AAAI} Symposium
  on Educational Advances in Artificial Intelligence, {EAAI} 2020, New York,
  NY, USA, February 7-12, 2020}, pages 8449--8456. {AAAI} Press.

\bibitem[{Ma et~al.(2021)Ma, Ilievski, Francis, Bisk, Nyberg, and
  Oltramari}]{ma2021knowledge}
Kaixin Ma, Filip Ilievski, Jonathan Francis, Yonatan Bisk, Eric Nyberg, and
  Alessandro Oltramari. 2021.
\newblock \href {https://ojs.aaai.org/index.php/AAAI/article/view/17593}
  {Knowledge-driven data construction for zero-shot evaluation in commonsense
  question answering}.
\newblock In \emph{Thirty-Fifth {AAAI} Conference on Artificial Intelligence,
  {AAAI} 2021, Thirty-Third Conference on Innovative Applications of Artificial
  Intelligence, {IAAI} 2021, The Eleventh Symposium on Educational Advances in
  Artificial Intelligence, {EAAI} 2021, Virtual Event, February 2-9, 2021},
  pages 13507--13515. {AAAI} Press.

\bibitem[{McCarthy(1959)}]{McCarthy_Programs59}
John McCarthy. 1959.
\newblock \href {http://www-formal.stanford.edu/jmc/mcc59.html} {Programs with
  common sense}.
\newblock In \emph{Proceedings of the {T}eddington Conference on the
  Mechanization of Thought Processes}, pages 75--91, London. Her Majesty's
  Stationary Office.

\bibitem[{Mihaylov et~al.(2018)Mihaylov, Clark, Khot, and
  Sabharwal}]{mihaylov-etal-2018-suit}
Todor Mihaylov, Peter Clark, Tushar Khot, and Ashish Sabharwal. 2018.
\newblock \href {https://doi.org/10.18653/v1/D18-1260} {Can a suit of armor
  conduct electricity? a new dataset for open book question answering}.
\newblock In \emph{Proceedings of the 2018 Conference on Empirical Methods in
  Natural Language Processing}, pages 2381--2391, Brussels, Belgium.
  Association for Computational Linguistics.

\bibitem[{Mitra et~al.(2019)Mitra, Banerjee, Pal, Mishra, and
  Baral}]{mitra2019additional}
Arindam Mitra, Pratyay Banerjee, Kuntal~Kumar Pal, Swaroop Mishra, and Chitta
  Baral. 2019.
\newblock \href {https://arxiv.org/abs/1909.08855} {How additional knowledge
  can improve natural language commonsense question answering?}
\newblock \emph{ArXiv preprint}, abs/1909.08855.

\bibitem[{Mostafazadeh et~al.(2016)Mostafazadeh, Chambers, He, Parikh, Batra,
  Vanderwende, Kohli, and Allen}]{mostafazadeh-etal-2016-corpus}
Nasrin Mostafazadeh, Nathanael Chambers, Xiaodong He, Devi Parikh, Dhruv Batra,
  Lucy Vanderwende, Pushmeet Kohli, and James Allen. 2016.
\newblock \href {https://doi.org/10.18653/v1/N16-1098} {A corpus and cloze
  evaluation for deeper understanding of commonsense stories}.
\newblock In \emph{Proceedings of the 2016 Conference of the North {A}merican
  Chapter of the Association for Computational Linguistics: Human Language
  Technologies}, pages 839--849, San Diego, California. Association for
  Computational Linguistics.

\bibitem[{Mostafazadeh et~al.(2020)Mostafazadeh, Kalyanpur, Moon, Buchanan,
  Berkowitz, Biran, and Chu-Carroll}]{mostafazadeh2020glucose}
Nasrin Mostafazadeh, Aditya Kalyanpur, Lori Moon, David Buchanan, Lauren
  Berkowitz, Or~Biran, and Jennifer Chu-Carroll. 2020.
\newblock Glucose: Generalized and contextualized story explanations.
\newblock In \emph{Proceedings of the 2020 Conference on Empirical Methods in
  Natural Language Processing (EMNLP)}, pages 4569--4586.

\bibitem[{OpenAI(2022)}]{chatgpt}
OpenAI. 2022.
\newblock \href {https://openai.com/blog/chatgpt} {Introducing chatgpt}.
\newblock 2022, Nov 30.

\bibitem[{Pallier et~al.(2011)Pallier, Devauchelle, and
  Dehaene}]{pallier2011cortical}
Christophe Pallier, Anne-Dominique Devauchelle, and Stanislas Dehaene. 2011.
\newblock Cortical representation of the constituent structure of sentences.
\newblock \emph{Proceedings of the National Academy of Sciences},
  108(6):2522--2527.

\bibitem[{Paranjape et~al.(2021)Paranjape, Michael, Ghazvininejad, Hajishirzi,
  and Zettlemoyer}]{paranjape-etal-2021-prompting}
Bhargavi Paranjape, Julian Michael, Marjan Ghazvininejad, Hannaneh Hajishirzi,
  and Luke Zettlemoyer. 2021.
\newblock \href {https://doi.org/10.18653/v1/2021.findings-acl.366} {Prompting
  contrastive explanations for commonsense reasoning tasks}.
\newblock In \emph{Findings of the Association for Computational Linguistics:
  ACL-IJCNLP 2021}, pages 4179--4192, Online. Association for Computational
  Linguistics.

\bibitem[{Peng et~al.(2022)Peng, Li, Wiegreffe, and Riedl}]{peng2021inferring}
Xiangyu Peng, Siyan Li, Sarah Wiegreffe, and Mark Riedl. 2022.
\newblock \href {https://aclanthology.org/2022.findings-emnlp.520} {Inferring
  the reader: Guiding automated story generation with commonsense reasoning}.
\newblock In \emph{Findings of the Association for Computational Linguistics:
  EMNLP 2022}, pages 7008--7029, Abu Dhabi, United Arab Emirates. Association
  for Computational Linguistics.

\bibitem[{Rajani et~al.(2019)Rajani, McCann, Xiong, and
  Socher}]{rajani-etal-2019-explain}
Nazneen~Fatema Rajani, Bryan McCann, Caiming Xiong, and Richard Socher. 2019.
\newblock \href {https://doi.org/10.18653/v1/P19-1487} {Explain yourself!
  leveraging language models for commonsense reasoning}.
\newblock In \emph{Proceedings of the 57th Annual Meeting of the Association
  for Computational Linguistics}, pages 4932--4942, Florence, Italy.
  Association for Computational Linguistics.

\bibitem[{Roemmele et~al.(2011)Roemmele, Bejan, and
  Gordon}]{roemmele2011choice}
Melissa Roemmele, Cosmin~Adrian Bejan, and Andrew~S Gordon. 2011.
\newblock Choice of plausible alternatives: An evaluation of commonsense causal
  reasoning.
\newblock In \emph{2011 AAAI Spring Symposium Series}.

\bibitem[{Sakaguchi et~al.(2020)Sakaguchi, Bras, Bhagavatula, and
  Choi}]{sakaguchi2021winogrande}
Keisuke Sakaguchi, Ronan~Le Bras, Chandra Bhagavatula, and Yejin Choi. 2020.
\newblock \href {https://aaai.org/ojs/index.php/AAAI/article/view/6399}
  {Winogrande: An adversarial winograd schema challenge at scale}.
\newblock In \emph{The Thirty-Fourth {AAAI} Conference on Artificial
  Intelligence, {AAAI} 2020, The Thirty-Second Innovative Applications of
  Artificial Intelligence Conference, {IAAI} 2020, The Tenth {AAAI} Symposium
  on Educational Advances in Artificial Intelligence, {EAAI} 2020, New York,
  NY, USA, February 7-12, 2020}, pages 8732--8740. {AAAI} Press.

\bibitem[{Sap et~al.(2019{\natexlab{a}})Sap, Bras, Allaway, Bhagavatula,
  Lourie, Rashkin, Roof, Smith, and Choi}]{sap2019atomic}
Maarten Sap, Ronan~Le Bras, Emily Allaway, Chandra Bhagavatula, Nicholas
  Lourie, Hannah Rashkin, Brendan Roof, Noah~A. Smith, and Yejin Choi.
  2019{\natexlab{a}}.
\newblock \href {https://doi.org/10.1609/aaai.v33i01.33013027} {{ATOMIC:} an
  atlas of machine commonsense for if-then reasoning}.
\newblock In \emph{The Thirty-Third {AAAI} Conference on Artificial
  Intelligence, {AAAI} 2019, The Thirty-First Innovative Applications of
  Artificial Intelligence Conference, {IAAI} 2019, The Ninth {AAAI} Symposium
  on Educational Advances in Artificial Intelligence, {EAAI} 2019, Honolulu,
  Hawaii, USA, January 27 - February 1, 2019}, pages 3027--3035. {AAAI} Press.

\bibitem[{Sap et~al.(2019{\natexlab{b}})Sap, Rashkin, Chen, Le~Bras, and
  Choi}]{sap-etal-2019-social}
Maarten Sap, Hannah Rashkin, Derek Chen, Ronan Le~Bras, and Yejin Choi.
  2019{\natexlab{b}}.
\newblock \href {https://doi.org/10.18653/v1/D19-1454} {Social {IQ}a:
  Commonsense reasoning about social interactions}.
\newblock In \emph{Proceedings of the 2019 Conference on Empirical Methods in
  Natural Language Processing and the 9th International Joint Conference on
  Natural Language Processing (EMNLP-IJCNLP)}, pages 4463--4473, Hong Kong,
  China. Association for Computational Linguistics.

\bibitem[{Schacter and Addis(2007)}]{schacter2007cognitive}
Daniel~L Schacter and Donna~Rose Addis. 2007.
\newblock The cognitive neuroscience of constructive memory: remembering the
  past and imagining the future.
\newblock \emph{Philosophical Transactions of the Royal Society B: Biological
  Sciences}, 362:773--786.

\bibitem[{Schank and Abelson(1977)}]{reason:SchAbe77a}
R.~C. Schank and R.~P. Abelson. 1977.
\newblock \emph{Scripts, Plans, Goals and Understanding: An Inquiry into Human
  Knowledge Structures}.
\newblock Lawrence Erlbaum, Hillsdale, NJ.

\bibitem[{Schank(1995)}]{schank1995tell}
R.C. Schank. 1995.
\newblock \href {https://books.google.co.jp/books?id=3fah9UGzVJ8C} {\emph{Tell
  Me a Story: Narrative and Intelligence}}.
\newblock Rethinking theory. Northwestern University Press.

\bibitem[{Schank(1983)}]{schank1983dynamic}
Roger~C Schank. 1983.
\newblock \emph{Dynamic memory: A theory of reminding and learning in computers
  and people}.
\newblock cambridge university press.

\bibitem[{Shi et~al.(2023)Shi, Wang, Fang, Xu, Ding, Liu, and
  Song}]{shi2023qadynamics}
Haochen Shi, Weiqi Wang, Tianqing Fang, Baixuan Xu, Wenxuan Ding, Xin Liu, and
  Yangqiu Song. 2023.
\newblock \href {https://arxiv.org/abs/2310.11303} {Qadynamics: Training
  dynamics-driven synthetic qa diagnostic for zero-shot commonsense question
  answering}.
\newblock \emph{ArXiv preprint}, abs/2310.11303.

\bibitem[{Shwartz and Choi(2020)}]{shwartz-choi-2020-neural}
Vered Shwartz and Yejin Choi. 2020.
\newblock \href {https://doi.org/10.18653/v1/2020.coling-main.605} {Do neural
  language models overcome reporting bias?}
\newblock In \emph{Proceedings of the 28th International Conference on
  Computational Linguistics}, pages 6863--6870, Barcelona, Spain (Online).
  International Committee on Computational Linguistics.

\bibitem[{Shwartz et~al.(2020)Shwartz, West, Le~Bras, Bhagavatula, and
  Choi}]{shwartz-etal-2020-unsupervised}
Vered Shwartz, Peter West, Ronan Le~Bras, Chandra Bhagavatula, and Yejin Choi.
  2020.
\newblock \href {https://doi.org/10.18653/v1/2020.emnlp-main.373} {Unsupervised
  commonsense question answering with self-talk}.
\newblock In \emph{Proceedings of the 2020 Conference on Empirical Methods in
  Natural Language Processing (EMNLP)}, pages 4615--4629, Online. Association
  for Computational Linguistics.

\bibitem[{Singh et~al.(2002)Singh, Lin, Mueller, Lim, Perkins, and
  Zhu}]{10.5555/646748.701499}
Push Singh, Thomas Lin, Erik~T. Mueller, Grace Lim, Travell Perkins, and Wan~Li
  Zhu. 2002.
\newblock Open mind common sense: Knowledge acquisition from the general
  public.
\newblock In \emph{On the Move to Meaningful Internet Systems, 2002 -
  DOA/CoopIS/ODBASE 2002 Confederated International Conferences DOA, CoopIS and
  ODBASE 2002}, page 1223–1237, Berlin, Heidelberg. Springer-Verlag.

\bibitem[{Singh et~al.(2021)Singh, Wen, Hou, Alipoormolabashi, Wu, Ma, and
  Peng}]{singh-etal-2021-com2sense}
Shikhar Singh, Nuan Wen, Yu~Hou, Pegah Alipoormolabashi, Te-lin Wu, Xuezhe Ma,
  and Nanyun Peng. 2021.
\newblock \href {https://doi.org/10.18653/v1/2021.findings-acl.78}
  {{COM}2{SENSE}: A commonsense reasoning benchmark with complementary
  sentences}.
\newblock In \emph{Findings of the Association for Computational Linguistics:
  ACL-IJCNLP 2021}, pages 883--898, Online. Association for Computational
  Linguistics.

\bibitem[{Speer et~al.(2017)Speer, Chin, and Havasi}]{speer2017conceptnet}
Robyn Speer, Joshua Chin, and Catherine Havasi. 2017.
\newblock \href {http://aaai.org/ocs/index.php/AAAI/AAAI17/paper/view/14972}
  {Conceptnet 5.5: An open multilingual graph of general knowledge}.
\newblock In \emph{Proceedings of the Thirty-First {AAAI} Conference on
  Artificial Intelligence, February 4-9, 2017, San Francisco, California,
  {USA}}, pages 4444--4451. {AAAI} Press.

\bibitem[{Srivastava et~al.(2022)Srivastava, Rastogi, Rao, Shoeb, Abid, Fisch,
  Brown, Santoro, Gupta, Garriga-Alonso et~al.}]{srivastava2022beyond}
Aarohi Srivastava, Abhinav Rastogi, Abhishek Rao, Abu Awal~Md Shoeb, Abubakar
  Abid, Adam Fisch, Adam~R Brown, Adam Santoro, Aditya Gupta, Adri{\`a}
  Garriga-Alonso, et~al. 2022.
\newblock \href {https://arxiv.org/abs/2206.04615} {Beyond the imitation game:
  Quantifying and extrapolating the capabilities of language models}.
\newblock \emph{ArXiv preprint}, abs/2206.04615.

\bibitem[{Tafjord et~al.(2019{\natexlab{a}})Tafjord, Clark, Gardner, Yih, and
  Sabharwal}]{tafjord2019quarel}
Oyvind Tafjord, Peter Clark, Matt Gardner, Wen{-}tau Yih, and Ashish Sabharwal.
  2019{\natexlab{a}}.
\newblock \href {https://doi.org/10.1609/aaai.v33i01.33017063} {{QUAREL:} {A}
  dataset and models for answering questions about qualitative relationships}.
\newblock In \emph{The Thirty-Third {AAAI} Conference on Artificial
  Intelligence, {AAAI} 2019, The Thirty-First Innovative Applications of
  Artificial Intelligence Conference, {IAAI} 2019, The Ninth {AAAI} Symposium
  on Educational Advances in Artificial Intelligence, {EAAI} 2019, Honolulu,
  Hawaii, USA, January 27 - February 1, 2019}, pages 7063--7071. {AAAI} Press.

\bibitem[{Tafjord et~al.(2019{\natexlab{b}})Tafjord, Gardner, Lin, and
  Clark}]{tafjord-etal-2019-quartz}
Oyvind Tafjord, Matt Gardner, Kevin Lin, and Peter Clark. 2019{\natexlab{b}}.
\newblock \href {https://doi.org/10.18653/v1/D19-1608} {{Q}ua{RT}z: An
  open-domain dataset of qualitative relationship questions}.
\newblock In \emph{Proceedings of the 2019 Conference on Empirical Methods in
  Natural Language Processing and the 9th International Joint Conference on
  Natural Language Processing (EMNLP-IJCNLP)}, pages 5941--5946, Hong Kong,
  China. Association for Computational Linguistics.

\bibitem[{Talmor et~al.(2019)Talmor, Herzig, Lourie, and
  Berant}]{talmor-etal-2019-commonsenseqa}
Alon Talmor, Jonathan Herzig, Nicholas Lourie, and Jonathan Berant. 2019.
\newblock \href {https://doi.org/10.18653/v1/N19-1421} {{C}ommonsense{QA}: A
  question answering challenge targeting commonsense knowledge}.
\newblock In \emph{Proceedings of the 2019 Conference of the North {A}merican
  Chapter of the Association for Computational Linguistics: Human Language
  Technologies, Volume 1 (Long and Short Papers)}, pages 4149--4158,
  Minneapolis, Minnesota. Association for Computational Linguistics.

\bibitem[{Talmor et~al.(2021)Talmor, Yoran, Le~Bras, Bhagavatula, Goldberg,
  Choi, and Berant}]{talmor2021commonsenseqa}
Alon Talmor, Ori Yoran, Ronan Le~Bras, Chandra Bhagavatula, Yoav Goldberg,
  Yejin Choi, and Jonathan Berant. 2021.
\newblock Commonsenseqa 2.0: Exposing the limits of ai through gamification.
\newblock In \emph{Thirty-fifth Conference on Neural Information Processing
  Systems Datasets and Benchmarks Track (Round 1)}.

\bibitem[{Taori et~al.(2023)Taori, Gulrajani, Zhang, Dubois, Li, Guestrin,
  Liang, and Hashimoto}]{alpaca}
Rohan Taori, Ishaan Gulrajani, Tianyi Zhang, Yann Dubois, Xuechen Li, Carlos
  Guestrin, Percy Liang, and Tatsunori~B. Hashimoto. 2023.
\newblock Stanford alpaca: An instruction-following llama model.
\newblock \url{https://github.com/tatsu-lab/stanford_alpaca}.

\bibitem[{Thoppilan et~al.(2022)Thoppilan, De~Freitas, Hall, Shazeer,
  Kulshreshtha, Cheng, Jin, Bos, Baker, Du et~al.}]{thoppilan2022lamda}
Romal Thoppilan, Daniel De~Freitas, Jamie Hall, Noam Shazeer, Apoorv
  Kulshreshtha, Heng-Tze Cheng, Alicia Jin, Taylor Bos, Leslie Baker, Yu~Du,
  et~al. 2022.
\newblock \href {https://arxiv.org/abs/2201.08239} {Lamda: Language models for
  dialog applications}.
\newblock \emph{ArXiv preprint}, abs/2201.08239.

\bibitem[{Touvron et~al.(2023)Touvron, Lavril, Izacard, Martinet, Lachaux,
  Lacroix, Rozi{\`e}re, Goyal, Hambro, Azhar et~al.}]{touvron2023llama}
Hugo Touvron, Thibaut Lavril, Gautier Izacard, Xavier Martinet, Marie-Anne
  Lachaux, Timoth{\'e}e Lacroix, Baptiste Rozi{\`e}re, Naman Goyal, Eric
  Hambro, Faisal Azhar, et~al. 2023.
\newblock \href {https://arxiv.org/abs/2302.13971} {Llama: Open and efficient
  foundation language models}.
\newblock \emph{ArXiv preprint}, abs/2302.13971.

\bibitem[{Tulving(2002)}]{tulving2002episodic}
Endel Tulving. 2002.
\newblock Episodic memory: From mind to brain.
\newblock \emph{Annual review of psychology}, 53:1--25.

\bibitem[{Wang et~al.(2019)Wang, Liang, Zhang, Li, and
  Gao}]{wang-etal-2019-make}
Cunxiang Wang, Shuailong Liang, Yue Zhang, Xiaonan Li, and Tian Gao. 2019.
\newblock \href {https://doi.org/10.18653/v1/P19-1393} {Does it make sense? and
  why? a pilot study for sense making and explanation}.
\newblock In \emph{Proceedings of the 57th Annual Meeting of the Association
  for Computational Linguistics}, pages 4020--4026, Florence, Italy.
  Association for Computational Linguistics.

\bibitem[{Wang et~al.(2023{\natexlab{a}})Wang, Sun, Chen, Li, and
  Gao}]{wang2023boosting}
Jianing Wang, Qiushi Sun, Nuo Chen, Xiang Li, and Ming Gao. 2023{\natexlab{a}}.
\newblock \href {https://arxiv.org/abs/2306.06427} {Boosting language models
  reasoning with chain-of-knowledge prompting}.
\newblock \emph{ArXiv preprint}, abs/2306.06427.

\bibitem[{Wang et~al.(2022{\natexlab{a}})Wang, Chan, Ilievski, Chen, and
  Ren}]{wang2022pinto}
PeiFeng Wang, Aaron Chan, Filip Ilievski, Muhao Chen, and Xiang Ren.
  2022{\natexlab{a}}.
\newblock Pinto: Faithful language reasoning using prompt-generated rationales.
\newblock In \emph{The Eleventh International Conference on Learning
  Representations}.

\bibitem[{Wang et~al.(2023{\natexlab{b}})Wang, Wang, Li, Gao, Yin, and
  Ren}]{wang2023scott}
Peifeng Wang, Zhengyang Wang, Zheng Li, Yifan Gao, Bing Yin, and Xiang Ren.
  2023{\natexlab{b}}.
\newblock \href {https://arxiv.org/abs/2305.01879} {Scott: Self-consistent
  chain-of-thought distillation}.
\newblock \emph{ArXiv preprint}, abs/2305.01879.

\bibitem[{Wang et~al.(2023{\natexlab{c}})Wang, Fang, Ding, Xu, Liu, Song, and
  Bosselut}]{wang2023car}
Weiqi Wang, Tianqing Fang, Wenxuan Ding, Baixuan Xu, Xin Liu, Yangqiu Song, and
  Antoine Bosselut. 2023{\natexlab{c}}.
\newblock \href {https://arxiv.org/abs/2305.14869} {Car:
  Conceptualization-augmented reasoner for zero-shot commonsense question
  answering}.
\newblock \emph{ArXiv preprint}, abs/2305.14869.

\bibitem[{Wang et~al.(2023{\natexlab{d}})Wang, Fang, Xu, Bo, Song, and
  Chen}]{wang-etal-2023-cat}
Weiqi Wang, Tianqing Fang, Baixuan Xu, Chun Yi~Louis Bo, Yangqiu Song, and Lei
  Chen. 2023{\natexlab{d}}.
\newblock \href {https://doi.org/10.18653/v1/2023.acl-long.733} {{CAT}: A
  contextualized conceptualization and instantiation framework for commonsense
  reasoning}.
\newblock In \emph{Proceedings of the 61st Annual Meeting of the Association
  for Computational Linguistics (Volume 1: Long Papers)}, pages 13111--13140,
  Toronto, Canada. Association for Computational Linguistics.

\bibitem[{Wang et~al.(2022{\natexlab{b}})Wang, Kordi, Mishra, Liu, Smith,
  Khashabi, and Hajishirzi}]{wang2022self}
Yizhong Wang, Yeganeh Kordi, Swaroop Mishra, Alisa Liu, Noah~A Smith, Daniel
  Khashabi, and Hannaneh Hajishirzi. 2022{\natexlab{b}}.
\newblock \href {https://arxiv.org/abs/2212.10560} {Self-instruct: Aligning
  language model with self generated instructions}.
\newblock \emph{ArXiv preprint}, abs/2212.10560.

\bibitem[{Wang et~al.(2023{\natexlab{e}})Wang, Zhang, Liang, and
  Li}]{wang-etal-2023-dynamic}
Yujie Wang, Hu~Zhang, Jiye Liang, and Ru~Li. 2023{\natexlab{e}}.
\newblock \href {https://doi.org/10.18653/v1/2023.acl-long.785} {Dynamic
  heterogeneous-graph reasoning with language models and knowledge
  representation learning for commonsense question answering}.
\newblock In \emph{Proceedings of the 61st Annual Meeting of the Association
  for Computational Linguistics (Volume 1: Long Papers)}, pages 14048--14063,
  Toronto, Canada. Association for Computational Linguistics.

\bibitem[{Wang et~al.(2023{\natexlab{f}})Wang, Do, Zhang, Zhang, Wang, Fang,
  Song, Wong, and See}]{wang-etal-2023-cola}
Zhaowei Wang, Quyet~V. Do, Hongming Zhang, Jiayao Zhang, Weiqi Wang, Tianqing
  Fang, Yangqiu Song, Ginny Wong, and Simon See. 2023{\natexlab{f}}.
\newblock \href {https://doi.org/10.18653/v1/2023.acl-long.288} {{COLA}:
  Contextualized commonsense causal reasoning from the causal inference
  perspective}.
\newblock In \emph{Proceedings of the 61st Annual Meeting of the Association
  for Computational Linguistics (Volume 1: Long Papers)}, pages 5253--5271,
  Toronto, Canada. Association for Computational Linguistics.

\bibitem[{Wei et~al.(2022)Wei, Wang, Schuurmans, Bosma, Xia, Chi, Le, Zhou
  et~al.}]{wei2022chain}
Jason Wei, Xuezhi Wang, Dale Schuurmans, Maarten Bosma, Fei Xia, Ed~Chi, Quoc~V
  Le, Denny Zhou, et~al. 2022.
\newblock Chain-of-thought prompting elicits reasoning in large language
  models.
\newblock \emph{Advances in Neural Information Processing Systems},
  35:24824--24837.

\bibitem[{Welbl et~al.(2017)Welbl, Liu, and
  Gardner}]{welbl-etal-2017-crowdsourcing}
Johannes Welbl, Nelson~F. Liu, and Matt Gardner. 2017.
\newblock \href {https://doi.org/10.18653/v1/W17-4413} {Crowdsourcing multiple
  choice science questions}.
\newblock In \emph{Proceedings of the 3rd Workshop on Noisy User-generated
  Text}, pages 94--106, Copenhagen, Denmark. Association for Computational
  Linguistics.

\bibitem[{Wen et~al.(2023)Wen, Tian, Wu, Yang, Shi, Huang, and
  Li}]{wen-etal-2023-grove}
Zhihua Wen, Zhiliang Tian, Wei Wu, Yuxin Yang, Yanqi Shi, Zhen Huang, and
  Dongsheng Li. 2023.
\newblock \href {https://doi.org/10.18653/v1/2023.findings-emnlp.262} {{GROVE}:
  A retrieval-augmented complex story generation framework with a forest of
  evidence}.
\newblock In \emph{Findings of the Association for Computational Linguistics:
  EMNLP 2023}, pages 3980--3998, Singapore. Association for Computational
  Linguistics.

\bibitem[{West et~al.(2022)West, Bhagavatula, Hessel, Hwang, Jiang, Le~Bras,
  Lu, Welleck, and Choi}]{west2021symbolic}
Peter West, Chandra Bhagavatula, Jack Hessel, Jena Hwang, Liwei Jiang, Ronan
  Le~Bras, Ximing Lu, Sean Welleck, and Yejin Choi. 2022.
\newblock \href {https://doi.org/10.18653/v1/2022.naacl-main.341} {Symbolic
  knowledge distillation: from general language models to commonsense models}.
\newblock In \emph{Proceedings of the 2022 Conference of the North American
  Chapter of the Association for Computational Linguistics: Human Language
  Technologies}, pages 4602--4625, Seattle, United States. Association for
  Computational Linguistics.

\bibitem[{Xie et~al.(2023)Xie, Cohn, and Lau}]{xie2023next}
Zhuohan Xie, Trevor Cohn, and Jey~Han Lau. 2023.
\newblock The next chapter: A study of large language models in storytelling.
\newblock In \emph{Proceedings of the 16th International Natural Language
  Generation Conference}, pages 323--351.

\bibitem[{Yao et~al.(2023)Yao, Wang, Mao, Tan, Huang, Chen, and
  Zhang}]{yao2023knowledge}
Yunzhi Yao, Peng Wang, Shengyu Mao, Chuanqi Tan, Fei Huang, Huajun Chen, and
  Ningyu Zhang. 2023.
\newblock \href {https://arxiv.org/abs/2305.08732} {Knowledge rumination for
  pre-trained language models}.
\newblock \emph{ArXiv preprint}, abs/2305.08732.

\bibitem[{Yasunaga et~al.(2021)Yasunaga, Ren, Bosselut, Liang, and
  Leskovec}]{yasunaga-etal-2021-qa}
Michihiro Yasunaga, Hongyu Ren, Antoine Bosselut, Percy Liang, and Jure
  Leskovec. 2021.
\newblock \href {https://doi.org/10.18653/v1/2021.naacl-main.45} {{QA}-{GNN}:
  Reasoning with language models and knowledge graphs for question answering}.
\newblock In \emph{Proceedings of the 2021 Conference of the North American
  Chapter of the Association for Computational Linguistics: Human Language
  Technologies}, pages 535--546, Online. Association for Computational
  Linguistics.

\bibitem[{Yu et~al.(2022)Yu, Iter, Wang, Xu, Ju, Sanyal, Zhu, Zeng, and
  Jiang}]{yu2022generate}
Wenhao Yu, Dan Iter, Shuohang Wang, Yichong Xu, Mingxuan Ju, Soumya Sanyal,
  Chenguang Zhu, Michael Zeng, and Meng Jiang. 2022.
\newblock Generate rather than retrieve: Large language models are strong
  context generators.
\newblock In \emph{The Eleventh International Conference on Learning
  Representations}.

\bibitem[{Zaccarella et~al.(2017)Zaccarella, Schell, and
  Friederici}]{zaccarella2017reviewing}
Emiliano Zaccarella, Marianne Schell, and Angela~D Friederici. 2017.
\newblock Reviewing the functional basis of the syntactic merge mechanism for
  language: A coordinate-based activation likelihood estimation meta-analysis.
\newblock \emph{Neuroscience \& Biobehavioral Reviews}, 80:646--656.

\bibitem[{Zellers et~al.(2018)Zellers, Bisk, Schwartz, and
  Choi}]{zellers-etal-2018-swag}
Rowan Zellers, Yonatan Bisk, Roy Schwartz, and Yejin Choi. 2018.
\newblock \href {https://doi.org/10.18653/v1/D18-1009} {{SWAG}: A large-scale
  adversarial dataset for grounded commonsense inference}.
\newblock In \emph{Proceedings of the 2018 Conference on Empirical Methods in
  Natural Language Processing}, pages 93--104, Brussels, Belgium. Association
  for Computational Linguistics.

\bibitem[{Zellers et~al.(2019)Zellers, Holtzman, Bisk, Farhadi, and
  Choi}]{zellers-etal-2019-hellaswag}
Rowan Zellers, Ari Holtzman, Yonatan Bisk, Ali Farhadi, and Yejin Choi. 2019.
\newblock \href {https://doi.org/10.18653/v1/P19-1472} {{H}ella{S}wag: Can a
  machine really finish your sentence?}
\newblock In \emph{Proceedings of the 57th Annual Meeting of the Association
  for Computational Linguistics}, pages 4791--4800, Florence, Italy.
  Association for Computational Linguistics.

\bibitem[{Zhang et~al.(2023{\natexlab{a}})Zhang, Xu, Wang, Li, Shi, Bi, and
  Lin}]{Zhang2023ASW}
Guangyao Zhang, Yangwen Xu, Xiuyi Wang, Jixing Li, W.~Shi, Yanchao Bi, and Nan
  Lin. 2023{\natexlab{a}}.
\newblock \href {https://api.semanticscholar.org/CorpusID:262124921} {A
  social-semantic working-memory account for two canonical language areas}.
\newblock \emph{Nature Human Behaviour}, 7:1980--1997.

\bibitem[{Zhang et~al.(2022)Zhang, Roller, Goyal, Artetxe, Chen, Chen, Dewan,
  Diab, Li, Lin et~al.}]{zhang2022opt}
Susan Zhang, Stephen Roller, Naman Goyal, Mikel Artetxe, Moya Chen, Shuohui
  Chen, Christopher Dewan, Mona Diab, Xian Li, Xi~Victoria Lin, et~al. 2022.
\newblock \href {https://arxiv.org/abs/2205.01068} {Opt: Open pre-trained
  transformer language models}.
\newblock \emph{ArXiv preprint}, abs/2205.01068.

\bibitem[{Zhang et~al.(2023{\natexlab{b}})Zhang, Storks, Hu, Sohn, Lee, Lee,
  and Chai}]{zhang2023heuristic}
Zheyuan Zhang, Shane Storks, Fengyuan Hu, Sungryull Sohn, Moontae Lee, Honglak
  Lee, and Joyce Chai. 2023{\natexlab{b}}.
\newblock \href {https://arxiv.org/abs/2310.18364} {From heuristic to analytic:
  Cognitively motivated strategies for coherent physical commonsense
  reasoning}.
\newblock \emph{ArXiv preprint}, abs/2310.18364.

\bibitem[{Zheng et~al.(2023)Zheng, Chiang, Sheng, Zhuang, Wu, Zhuang, Lin, Li,
  Li, Xing, Zhang, Gonzalez, and Stoica}]{zheng2023judging}
Lianmin Zheng, Wei-Lin Chiang, Ying Sheng, Siyuan Zhuang, Zhanghao Wu, Yonghao
  Zhuang, Zi~Lin, Zhuohan Li, Dacheng Li, Eric.~P Xing, Hao Zhang, Joseph~E.
  Gonzalez, and Ion Stoica. 2023.
\newblock \href {https://arxiv.org/abs/2306.05685} {Judging llm-as-a-judge with
  mt-bench and chatbot arena}.
\newblock \emph{ArXiv preprint}, abs/2306.05685.

\bibitem[{Zhong et~al.(2019)Zhong, Tang, Duan, Zhou, Wang, and
  Yin}]{zhong2019improving}
Wanjun Zhong, Duyu Tang, Nan Duan, Ming Zhou, Jiahai Wang, and Jian Yin. 2019.
\newblock Improving question answering by commonsense-based pre-training.
\newblock In \emph{Natural Language Processing and Chinese Computing: 8th CCF
  International Conference, NLPCC 2019, Dunhuang, China, October 9--14, 2019,
  Proceedings, Part I 8}, pages 16--28. Springer.

\bibitem[{Zhou et~al.(2022)Zhou, Gopalakrishnan, Hedayatnia, Kim, Pujara, Ren,
  Liu, and Hakkani-Tur}]{zhou-etal-2022-think}
Pei Zhou, Karthik Gopalakrishnan, Behnam Hedayatnia, Seokhwan Kim, Jay Pujara,
  Xiang Ren, Yang Liu, and Dilek Hakkani-Tur. 2022.
\newblock \href {https://doi.org/10.18653/v1/2022.acl-long.88} {Think before
  you speak: Explicitly generating implicit commonsense knowledge for response
  generation}.
\newblock In \emph{Proceedings of the 60th Annual Meeting of the Association
  for Computational Linguistics (Volume 1: Long Papers)}, pages 1237--1252,
  Dublin, Ireland. Association for Computational Linguistics.

\end{thebibliography}

\appendix

\section{Prompts} \label{prompts}

The prompts we use in this paper for instructing LLMs to generate stories and rules are shown in Table \ref{t1}. The common prefix which contains the question and the answer options (if applicable) is added before the two prompts below. To avoid responses such as ``As an AI language model, I do not have past experiences'', the specific name ``Jane'' is used instead of ``you'' in our prompts. It is worth noting that the choice of the name is arbitrary, and any name can be used. The potential influence of name biases in LLMs is a topic for future study.

The prompts for answering questions are shown in Table \ref{t1qa}. The prompt for automatically evaluating the stories and rules is shown in Table \ref{t1eval}.

These prompts are constructed through a prompt engineering process. This involves testing and comparing different prompt variations to select the most effective ones. Through this process, we can find the prompts to ensure optimal performance in guiding LLMs to generate high-quality stories and rules and to answer questions effectively.

\begin{table}[!t]
\centering
\small
\begin{tabular}{m{200pt}}
\hline
Common Prefix:\\
\textit{Jane is answering this question: }\\
\textit{Question: Where do adults use glue sticks? }\\
\textit{Options: A. classroom, B. desk drawer, C. at school, D. office, E. kitchen drawer. }\\
\hline
Prompt for Generating Story:\\
\textit{Jane is reminded of a specific past experience analogous to the situation and the most important information in this question. However, Jane refrains from forming conclusions or making guesses about the answer at this time.} \\
\textit{Write a possible experience as detailed and focused story in a paragraph that Jane may recall and conforms to the common practise. Do not use names in the question or mention the options in the story. Do not output extra sentences.} \\
\hline
Prompt for Generating Rule:\\
\textit{Jane is reminded of specific commonsense rules relevant to the situation and the most important information in this question (without considering the options). However, Jane refrains from forming conclusions or making guesses about the answer at this time.} \\
\textit{List possible commonsense rules as simple knowledge sentences that Jane may recall in a paragraph. Do not output extra sentences.} \\
\hline
\end{tabular}
\caption{\label{t1}
Prompts for generating stories and rules, with an example question from the CommonsenseQA dataset. The common prefix comes before each prompt.
}
\end{table}

\begin{table}[!t]
\centering
\small
\begin{tabular}{m{200pt}}
\hline
Prompt for Answering Question without Context:\\
\textit{Choose the most suitable answer for the question by selecting the answer letter and do not say anything else: \{question\} \{answer\_options\} } \\
\hline
Prompt for Answering Question with Story:\\
\textit{Read these experiences:} \\
\textit{\{story\}} \\
\textit{Analogy to the above text as reference, choose the most suitable answer for the question by selecting the answer letter and do not include anything else: \{question\} \{answer\_options\}} \\
\hline
Prompt for Answering Question with Rule:\\
\textit{Read these commonsense rules:} \\
\textit{\{rule\}} \\
\textit{Based on the above text as reference, choose the most suitable answer for the question by selecting the answer letter and do not include anything else: \{question\} \{answer\_options\}} \\
\hline
Prompt for Answering Question with Both Rule and Story:\\
\textit{Read these experiences:}\\
\textit{\{story\}} \\
\textit{Read these commonsense rules:} \\
\textit{\{rule\}} \\
\textit{Based on the above experiences and commonsense rules as reference, choose the most suitable answer for the question by selecting the answer letter and do not include the option content or anything else: \{question\} \{answer\_options\}} \\
\hline
\end{tabular}
\caption{\label{t1qa}
Prompts for answering questions with stories or rules. 
}
\end{table}

\begin{table}[!t]
\centering
\small
\begin{tabular}{m{200pt}}
\hline
Prompt for Evaluating Commonsense Accuracy:\\
\textit{Please evaluate the following sentences for common sense based on your commonsense knowledge:}\\
\textit{\{text\}}\\
\textit{Does the sentences align with your common sense? Respond with "yes" or "no" only.} \\
\hline
\end{tabular}
\caption{\label{t1eval}
Prompt for evaluating the stories and rules with ChatGPT.
}
\end{table}

\section{Details of Commonsense QA Datasets}
Table \ref{t5} provides information on the number of questions, question types, and accuracy when randomly selecting an answer option for each commonsense QA dataset. All these datasets are in English. It is important to note that, for HellaSWAG-act.net, HellaSWAG-wikihow, SWAG, and PROST datasets, we randomly sample 1,000 questions from each of their development sets. This decision is made due to the excessively large number of questions in their development sets, which would have required significant time and computational resources for evaluation (3,243 for HellaSWAG-act.net, 6,799 for HellaSWAG-wikihow, 20,006 for SWAG, and 18,736 for PROST). An example from each dataset is presented in Table \ref{t6}.

\begin{table*}[!t]
\centering
\small
\begin{tabular}{lrcc}
\hline
Dataset           & \#Questions          & Type                 & Random Accuracy \\ \hline
CommonsenseQA     & 1,221                & General              & 20.00\%         \\
OpenBookQA        & 500                  & Science              & 25.00\%         \\
PIQA              & 1,838                & General              & 50.00\%         \\
SocialIQA         & 1,954                & Social               & 33.33\%         \\
ARC-easy         & 570                  & Science              & 25.00\%         \\
ARC-challenge    & 299                  & Science              & 25.00\%         \\
QASC              & 926                  & Science              & 12.50\%         \\
AI2Sci-elem      & 123                  & Science              & 25.23\%         \\
AI2Sci-middle    & 125                  & Science              & 24.92\%         \\
WinoGrande        & 1,267                & General              & 50.00\%         \\
WSC               & 285                  & General              & 50.00\%         \\
NumerSense        & 200                  & Number               & 0.00\%          \\
HellaSWAG-act.net & 1,000                & Daily event          & 25.00\%         \\
HellaSWAG-wikihow & 1,000                & Daily event          & 25.00\%         \\
CommonsenseQA2.0  & 2,541                & Yes/No               & 50.00\%         \\
SWAG              & 1,000                & Daily event          & 25.00\%         \\
Com2Sense         & 782                  & Yes/No               & 50.00\%         \\
SciQ              & 1,000                & Science              & 25.00\%         \\
QuaRel            & 278                  & Science \& Comparing & 50.00\%         \\
QuaRTz            & 384                  & Science \& Comparing & 50.00\%         \\
CycIC             & 1,525                & Logical Reasoning    & 32.16\%         \\
ComVE (Task A)    & 997                  & Yes/No               & 50.00\%         \\
COPA              & 500                  & General              & 50.00\%         \\
PROST             & 1,000                & Physical             & 25.00\%         \\
CODAH             & 556                  & General              & 25.00\%         \\
SCT               & 1,571                & Daily event          & 50.00\%         \\
$\alpha$NLI       & 1,532                & Daily event          & 50.00\%         \\
WinoVenti         & 4,352                & General              & 50.00\%         \\ \hline
\end{tabular}
\caption{\label{t5}
Commonsense QA datasets used in this paper. Random accuracy means the accuracy of randomly choosing an answer option.
}
\end{table*}

\begin{table*}[!p]
\centering
\small
\resizebox{\linewidth}{!}{
\begin{tabular}{lm{395pt}}
\hline
Dataset           & Example                                                                                                                                                                                                                                                                                                                                                                                                                                                                                                                                                                                                                                                                                                                                                                                                                                                                                                                                                                                                                                                                                                                 \\ \hline
CommonsenseQA     & What is another name for a disk for storing information? A. computer store B. computer to store data \textbf{C. computer hard drive} D. cd player E. usb mouse                                                                                                                                                                                                                                                                                                                                                                                                                                                                                                                                                                                                                                                                                                                                                                                                                                                                                                                                                                   \\ \hline
OpenBookQA        & Owls spend their nights A. tending to their homes B. sleeping in hollow logs \textbf{C. scanning their territory for field mice} D. hanging out with other owls                                                                                                                                                                                                                                                                                                                                                                                                                                                                                                                                                                                                                                                                                                                                                                                                                                                                                                                                                                  \\ \hline
PIQA              & Where can I buy a tennis ball \textbf{A. You can purchase a tennis ball at any sports store} B. You can purchase a tennis racket at any sports store                                                                                                                                                                                                                                                                                                                                                                                                                                                                                                                                                                                                                                                                                                                                                                                                                                                                                                                                                                             \\ \hline
SocialIQA         & Aubrey the officer pulled a driver over for speeding on the road. Why did Aubrey do this? A. find a safe place to pull the person over \textbf{B. so people don't drive to fast} C. look up the person's license plate number                                                                                                                                                                                                                                                                                                                                                                                                                                                                                                                                                                                                                                                                                                                                                                                                                                                                                                    \\ \hline
ARC-easy         & Scientists at a local university have been studying the impact that people have on Earth. One of the areas being studied is how the burning of fossil fuels affects the environment. Which effect of fossil fuel burning have the scientists most likely evaluated? A. the production of nitrogen-fixing bacteria B. the mechanical weathering of roads \textbf{C. the formation of acid rain} D. the increase in runoff                                                                                                                                                                                                                                                                                                                                                                                                                                                                                                                                                                                                                                                                                                         \\ \hline
ARC-challenge    & How should a line graph be used to display distance and time data for a moving object? A. The y-axis should be labeled as time, which is the dependent variable. B. The y-axis should be labeled as distance, which is the independent variable. C. The x-axis should be labeled as distance, which is the dependent variable. \textbf{D. The x-axis should be labeled as time, which is the independent variable.}                                                                                                                                                                                                                                                                                                                                                                                                                                                                                                                                                                                                                                                                                                              \\ \hline
QASC              & What may renal failure be treated with? A. Laboratory B. Lymphocytes C. saves lives \textbf{D. dialysis} E. Lymph fluid F. dandelions G. ibuprofen H. Protein                                                                                                                                                                                                                                                                                                                                                                                                                                                                                                                                                                                                                                                                                                                                                                                                                                                                                                                                                                    \\ \hline
AI2Sci-elem      & To make an electromagnet, a conductor should be coiled around - A. a glass tube \textbf{B. an iron nail} C. a roll of paper D. a wooden stick                                                                                                                                                                                                                                                                                                                                                                                                                                                                                                                                                                                                                                                                                                                                                                                                                                                                                                                                                                                    \\ \hline
AI2Sci-middle    & Which best describes the characteristics of a river basin? \textbf{A. the land drained by a river and its tributaries} B. the land formed when rivers create estuaries and marshes C. the land at the mouth of a river where water flows into the ocean D. the land formed as a result of a river flooding                                                                                                                                                                                                                                                                                                                                                                                                                                                                                                                                                                                                                                                                                                                                                                                                                       \\ \hline
WinoGrande        & She chose the black car over the green car, because the A. black car has more brighter color. \textbf{B. green car has more brighter color.}                                                                                                                                                                                                                                                                                                                                                                                                                                                                                                                                                                                                                                                                                                                                                                                                                                                                                                                                                                                     \\ \hline
WSC               & The user changed his password from "GrWQWu8JyC" to "willow-towered Canopy Huntertropic wrestles" as A. grwqwu8jyc was easy to remember. \textbf{B. willow-towered canopy huntertropic wrestles was easy to remember.}                                                                                                                                                                                                                                                                                                                                                                                                                                                                                                                                                                                                                                                                                                                                                                                                                                                                                                            \\ \hline
NumerSense        & a french horn has \textless{}how many\textgreater\ keys. (\textbf{three})                                                                                                                                                                                                                                                                                                                                                                                                                                                                                                                                                                                                                                                                                                                                                                                                                                                                                                                                                                                                                                                         \\ \hline
HellaSWAG-act.net & Another man practices hurling himself backward over a pole onto a gym mat inside of the gym. several more men A. practice hurling street hurling outside and on a gym floor. B. practice hurling while a coach's hand watches. \textbf{C. practice long jumps and backward jumps inside of the gym using the sandbox and gym mats as landing tools.} D. practice pitches outside of the gym interior.                                                                                                                                                                                                                                                                                                                                                                                                                                                                                                                                                                                                                                                                                                                            \\ \hline
HellaSWAG-wikihow & {[}header{]} How to do tiger eye hair {[}title{]} Purchase your hair dye. {[}step{]} Take a trip to your local drug store or beauty supply store. Depending on the look you're going for, you may want to keep it simple and just choose one color, or you may want to buy four. A. Your hair will stick out more if you use a thicker dye, such as a mousse or gel. {[}title{]} Pour 3 ounces of red wine into your bowl. B. {[}substeps{]} In the tattoo artist's shop you should find several different colors and strips, and apply those to your hair. Make sure the colors match what you want to do. \textbf{C. It's totally up to you! You can buy a blonde highlighting kit that will lighten pieces of your brown hair, auburn dye, gorgeous golden hues, soft brown dyes-whichever dye you think will look good in your hair. The darker your hair is, the less of an effect you'll notice.} D. {[}substeps{]} If you want to dye your hair yourself, make sure to use the formula before you apply the dye. Gels are sometimes recommended but are highly expensive, and can be difficult to find at grocery stores. \\ \hline
CommonsenseQA2.0  & In the US a senator is a person elected to a six year term? \textbf{A. yes} B. no                                                                                                                                                                                                                                                                                                                                                                                                                                                                                                                                                                                                                                                                                                                                                                                                                                                                                                                                                                                                                                                \\ \hline
SWAG              & Someone finds people playing chess at one of the long polished tables. She \textbf{A. walks down the brightly decorated hall to join them.} B. pats him in the gut with the box. C. faces the building, in wonder, someone and the other recruits stand around watching, uneasy. D. approaches two fat men wearing an earpiece into an office.                                                                                                                                                                                                                                                                                                                                                                                                                                                                                                                                                                                                                                                                                                                                                                                   \\ \hline
Com2Sense         & Because the drive was 20 miles long, Beth was able to make it to her destination in under 5 minutes. A. True \textbf{B. False}                                                                                                                                                                                                                                                                                                                                                                                                                                                                                                                                                                                                                                                                                                                                                                                                                                                                                                                                                                                                   \\ \hline
SciQ              & What parts of a human possess the highest concentration of thermoreceptors? A. face and hair B. hand and ears \textbf{C. face and ears} D. hands and feet                                                                                                                                                                                                                                                                                                                                                                                                                                                                                                                                                                                                                                                                                                                                                                                                                                                                                                                                                                        \\ \hline
QuaRel            & Lebron James a strong player for the Cavs battles Kevin Durant a thin player for a rebound. Who is likely to get the rebound? A. Durant \textbf{B. Lebron}                                                                                                                                                                                                                                                                                                                                                                                                                                                                                                                                                                                                                                                                                                                                                                                                                                                                                                                                                                       \\ \hline
QuaRTz            & Long ago the surface of Venus warmed enough that greenhouse gases escaped into the atmosphere. As a result, the greenhouse effect on that planet \textbf{A. increased} B. decreased                                                                                                                                                                                                                                                                                                                                                                                                                                                                                                                                                                                                                                                                                                                                                                                                                                                                                                                                              \\ \hline
CycIC             & Rob lauded Will. Charity chastised Will. Who made Will feel happy? A. Daisy B. Cliff \textbf{C. Rob} D. Charity E. Joy                                                                                                                                                                                                                                                                                                                                                                                                                                                                                                                                                                                                                                                                                                                                                                                                                                                                                                                                                                                                           \\ \hline
ComVE (Task A)    & Which statement of the two is against common sense? \textbf{A. The cleaner is in charge of the money at the store} B. The cashier is in charge of the money at the store                                                                                                                                                                                                                                                                                                                                                                                                                                                                                                                                                                                                                                                                                                                                                                                                                                                                                                                                                         \\ \hline
COPA              & The woman was in a bad mood. What was the effect of this? A. She engaged in small talk with her friend. \textbf{B. She told her friend to leave her alone.}                                                                                                                                                                                                                                                                                                                                                                                                                                                                                                                                                                                                                                                                                                                                                                                                                                                                                                                                                                      \\ \hline
PROST             & A person drops a bottle, a mirror, an egg, and a shirt from a balcony. Which object is the least likely to break? A. bottle B. mirror C. egg \textbf{D. shirt}                                                                                                                                                                                                                                                                                                                                                                                                                                                                                                                                                                                                                                                                                                                                                                                                                                                                                                                                                                   \\ \hline
CODAH             & Kieran is a whale. Kieran \textbf{A. is a mammal} B. is a dog C. has six human kids D. is a orange                                                                                                                                                                                                                                                                                                                                                                                                                                                                                                                                                                                                                                                                                                                                                                                                                                                                                                                                                                                                                               \\ \hline
SCT               & I wanted to buy a video game console. I asked my parents, and they came up with an idea. They said if I did my chores, I would be given money to save. I did my chores without being asked every week for a whole summer. What is the end of this story? \textbf{A. My parents gave me enough money to buy the console.} B. At the end of the summer I gave the money back to my parents.                                                                                                                                                                                                                                                                                                                                                                                                                                                                                                                                                                                                                                                                                                                                        \\ \hline
$\alpha$NLI       & The beginning of the story: Jim got ready for his first date.The ending of the story: Since then, she has ignored all of Jim's text messages. What happened between the begining and the end of the story? \textbf{A. Jim's date wasn't attracted to him.} B. Jim went on the date and said he didn't like the girl.                                                                                                                                                                                                                                                                                                                                                                                                                                                                                                                                                                                                                                                                                                                                                                                                             \\ \hline
WinoVenti         & The walnut was painted. The walnut is A. edible \textbf{B. toxic}                                                                                                                                                                                                                                                                                                                                                                                                                                                                                                                                                                                                                                                                                                                                                                                                                                                                                                                                                                                                                                                                \\ \hline
\end{tabular}}
\caption{\label{t6}
Example question for each commonsense QA dataset. Answers are shown in bold.
}
\end{table*}

\begin{table}[!h]
\centering
\small
\begin{tabular}{c|ccc}
\hline
Setting & ChatGPT & Vicuna & Alpaca \\
\hline
Story & \textbf{96\%} & \textbf{94\%} & \textbf{88\%} \\
Rule & 90\% & 92\% & 78\% \\
\hline
\end{tabular}
\caption{\label{t-manual}
Manually evaluated commonsense accuracy of stories and rules.
}
\end{table}

\section{Further Analyses of Generated Stories} \label{sft-further}

\subsection{Manual Evaluations of Commonsense Accuracy in Generated Stories and Rules} \label{manual}

To further support the commonsense accuracy results presented in Section \ref{commonsense_acc}, we conduct a manual evaluation of the generated stories and rules. For each model, we sample 50 stories and 50 rules from the Commonsense QA dataset and manually label each one for commonsense correctness. The annotators are the authors of this paper. A story or rule is marked as incorrect if it contains a description that contradicts real-world commonsense. The results, shown in Table \ref{t-manual}, indicate that the commonsense accuracy of stories is higher than that of rules across all three models. This further supports our findings in Section \ref{commonsense_acc}, where ChatGPT is used to automatically assess the commonsense accuracy of a larger set of stories and rules.

\subsection{Accuracy Differences between Using Stories and Rules for ChatGPT and Alpaca} \label{acc-change}
\begin{figure}[!ht]
  \centering
  \includegraphics[width=\columnwidth]{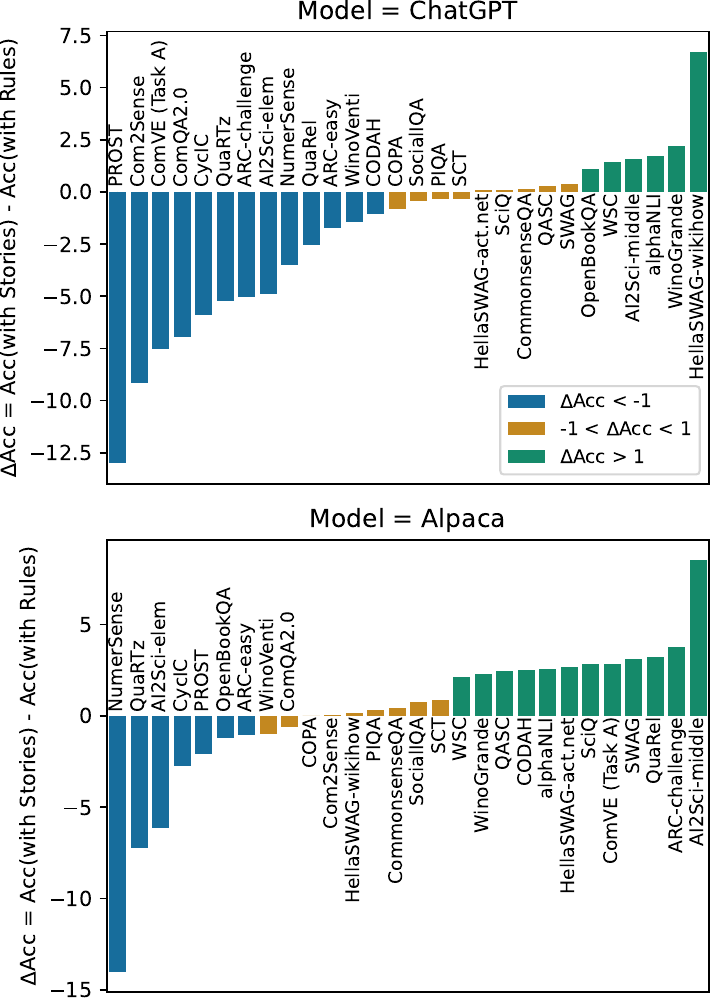}
  \caption{Comparison between the accuracy (\%) with stories and with rules for ChatGPT and Alpaca.}
  \label{f-acc-sup}
\end{figure}

We show the accuracy differences between using stories and rules for the ChatGPT and Alpaca models in Figure \ref{f-acc-sup}, as supplementary to Figure \ref{f-acc}.

\subsection{Error Analysis} \label{error_chart}

\begin{figure}[!ht]
\setlength{\belowcaptionskip}{0.3cm}
  \centering
  \includegraphics[width=0.95\columnwidth]{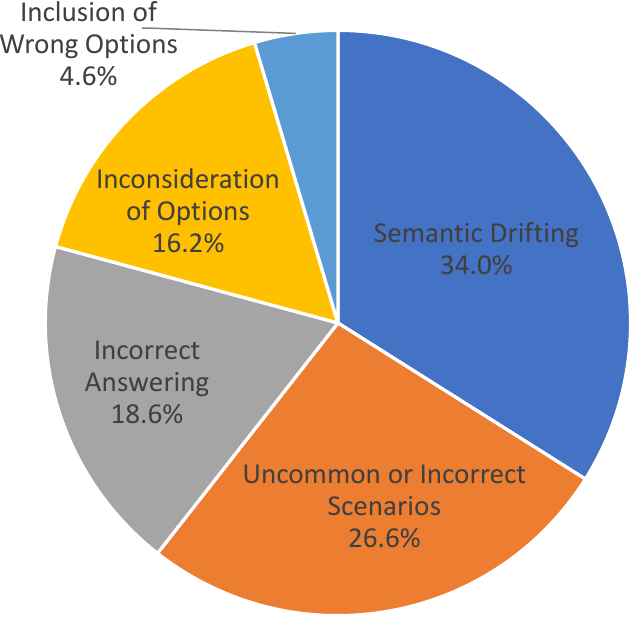}
  \caption{Error analysis of stories generated by Vicuna.}
  \label{f3}
\end{figure}

We show the pie chart of our error analysis in Figure \ref{f3}. From the figure we can see that uncommon or incorrect scenarios and semantic drifting are the primary reasons for incorrect model answers, accounting for over 60\% of total errors.

\subsection{Correlation Between Answer Accuracy and the Two Scores of Stories} \label{correlation}

\begin{figure}[!t]
  \centering
  \includegraphics[width=0.98\columnwidth]{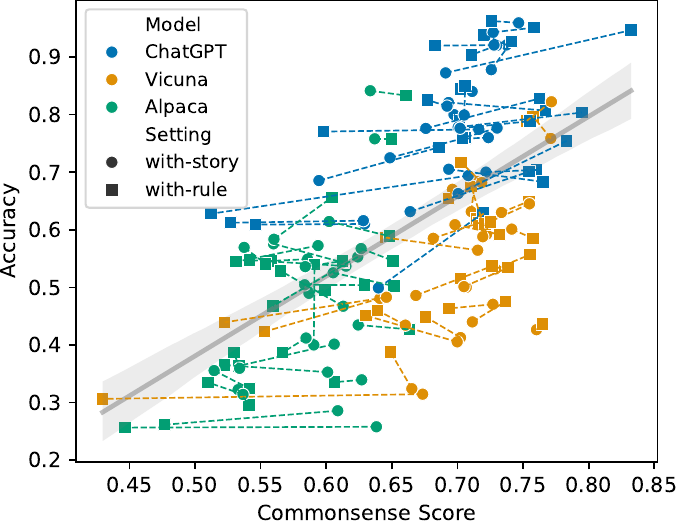}
  \caption{Correlation between answer accuracy of each dataset and commonsense scores of stories and rules.}
  \label{f5}
\end{figure}

\begin{figure}[!t]
  \centering
  \includegraphics[width=0.95\columnwidth]{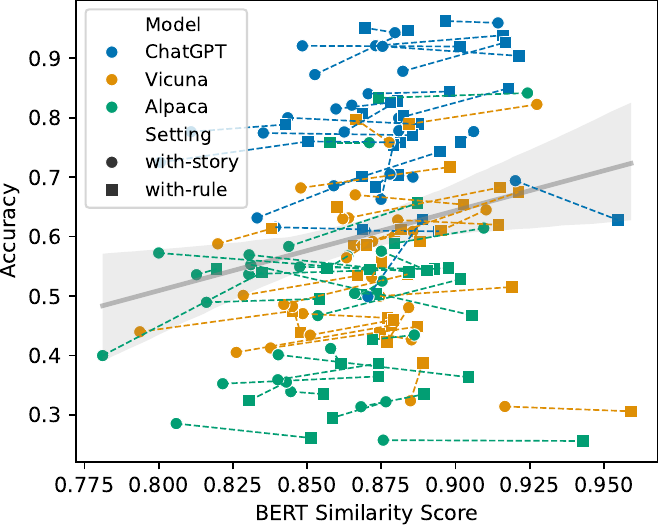}
  \caption{Correlation between answer accuracy of each dataset and BERT similarity scores of stories and rules.}
  \label{f6}
\end{figure}

We plot the correlation between answer accuracy and the commonsense score in Figure \ref{f5}, and the BERT similarity score in Figure \ref{f6}. The dashed lines are connected between the results of using stories and rules of the same datasets, which further reveal positive correlations of the two scores between using stories and using rules on most datasets

\subsection{Length of Stories and Rules}

We calculate the average lengths of a single story and rule generated by the models used in our study. Lengths are calculated as the number of words while excluding stop words. The results are shown in Table \ref{t_len}. This analysis reveals that, on average, stories are longer than rules. However, this difference in length does not imply a difference in the amount of information conveyed. Stories tend to include longer sentences with detailed descriptions and personal feelings, incorporating adjectives and adverbs, while rules are typically more concise and straightforward. Despite these stylistic differences, both formats provide a comparable amount of commonsense information.

\begin{table}[!t]
\setlength{\belowcaptionskip}{-0.1cm}
\centering
\small
\begin{tabular}{ccc}
\hline
Model & Story Length & Rule Length \\
\hline
ChatGPT & 82.6 & 65.5 \\
Alpaca & 50.1 & 40.7 \\
Vicuna & 62.2 & 49.9 \\
\hline
\end{tabular}
\caption{\label{t_len}
Average lengths (number of words) of generated stories and rules, excluding stop words.
}
\end{table}

\section{Hyper-parameters for Self-SFT} \label{parameter}

For the filtering step, we randomly select 200 questions initially answered incorrectly by the Vicuna and can be corrected with at least one generated story for each dataset. The filter ratio $K$ is set to 50\%, resulting in 2,269, 2,300, and 2,447 training stories across the 8 datasets for three iterations.

For the training step, we use LoRA tuning \cite{hu2021lora} for training the Vicuna model. LoRA is a parameter-efficient fine-tuning method that has become a common practice in the LLM era to reduce the overhead of expensive adaptations. The hyper-parameters for LoRA are rank $r = 16$ and $\alpha = 16$. Models are fine-tuned for 3 epochs in each iteration with a batch size of 64 and a learning rate of 3e-4. 

\section{More Results and the Ablation Study of Self-SFT} \label{ablation}

\subsection{Full Accuracy on All Datasets}

We show the accuracy before and after self-SFT of each dataset in Table \ref{t7}. Across 22 datasets, our self-SFT method consistently outperforms the original Vicuna model without SFT (7 seen and 15 unseen) and shows the most accuracy improvements at iterations 1 and 2.

\subsection{Effect of Iterative Self-SFT}

We compare our self-SFT method with a naive SFT method without iteration and scoring. In this ablated method, we only filter the stories generated by Vicuna on the training sets that can rectify an initially incorrect answer to correct, and use these stories for fine-tuning. The results are shown in Table \ref{t7}.

From the results, we can see that the ablated method performs worse than our self-SFT method on all datasets, and even worse than the original Vicuna model without fine-tuning on some datasets. This further verifies that our iterative scoring and filtering mechanism is crucial for alleviating the semantic drifting and commonsense hallucination issues of story generation.

\begin{table*}[!t]
\setlength{\belowcaptionskip}{-0.1cm}
\centering
\small
\begin{tabular}{l|c|ccc|c}
\hline
Dataset                    & No SFT         & Iter-1         & Iter-2         & Iter-3         & Ablation \\ \hline
\textbf{HellaSWAG-act.net} & 48.07          & 46.63          & \textbf{48.49} & 46.19          & 47.40    \\
\textbf{SWAG}              & 48.28          & \textbf{48.38} & 47.88          & 47.03          & 47.70    \\
\textbf{$\alpha$NLI}       & 64.51          & 65.47          & 65.99          & \textbf{66.51} & 64.34    \\
SCT                        & 82.24          & 82.78          & \textbf{82.80} & 82.40          & 82.50    \\
QuaRel                     & \textbf{60.89} & 59.93          & 59.42          & 60.73          & 59.23    \\
PIQA                       & 67.03          & \textbf{69.07} & 68.04          & 68.24          & 67.90    \\
\textbf{WinoGrande}        & 60.11          & 61.42          & \textbf{61.83} & 60.10          & 58.80    \\
\textbf{AI2Sci-elem}      & 62.81          & 62.60          & \textbf{65.57} & 64.23          & 60.16    \\
\textbf{HellaSWAG-wikihow} & \textbf{31.44} & 30.65          & 30.97          & 29.34          & 30.90    \\
CommonsenseQA              & \textbf{47.05} & 44.99          & 45.07          & 43.74          & 43.57    \\
SocialIQA                  & 42.64          & 41.59          & \textbf{43.45} & 41.64          & 42.58    \\
ARC-challenge             & 50.00          & \textbf{50.84} & 47.32          & 48.66          & 46.49    \\
WSC                        & 63.00          & \textbf{65.60} & 64.79          & 63.96          & 63.16    \\
CODAH                      & 56.45          & 58.04          & 57.43          & \textbf{58.76} & 56.83    \\
Com2Sense                  & \textbf{53.20} & 51.15          & 52.17          & 52.69          & 51.48    \\
WinoVenti                  & 58.79          & \textbf{59.80} & 59.39          & 58.90          & 58.32    \\
CycIC                      & 43.44          & \textbf{44.39} & 39.42          & 40.32          & 42.95    \\
QuaRTz                     & 58.52          & 59.53          & \textbf{60.63} & 56.69          & 59.11    \\
CommonsenseQA2.0           & 50.12          & \textbf{51.15} & 50.22          & 50.79          & 48.92    \\
NumerSense                 & \textbf{44.00} & 40.50          & 43.50          & 43.00          & 43.00    \\
SciQ                       & 68.19          & 68.21          & 68.47          & \textbf{70.87} & 67.80    \\
\textbf{OpenBookQA}        & 41.28          & 43.43          & 42.11          & \textbf{44.33} & 43.20    \\
COPA                       & 75.86          & \textbf{77.08} & 77.06          & 76.75          & 74.80    \\
QASC                       & 40.53          & 39.72          & 39.33          & \textbf{40.70} & 36.61    \\
ComVE (Task A)             & 48.59          & \textbf{49.15} & 48.35          & 48.14          & 48.55    \\
ARC-easy                  & \textbf{63.15} & 62.46          & 62.21          & 61.97          & 61.75    \\
PROST                      & 32.40          & \textbf{35.46} & 34.04          & 32.62          & 31.00    \\
\textbf{AI2Sci-middle}    & 59.20          & 60.00          & \textbf{63.20} & 59.20          & 58.40    \\
\hline
\end{tabular}

\caption{\label{t7}
Accuracy (\%) of commonsense QA by Vicuna with and without iterative Self-SFT and with ablation study. Bold dataset names are seen datasets during self-SFT.
}
\end{table*}

\subsection{Effect of Hyper-parameter: Filter Ratio}

\begin{figure}[!t]
  \centering
  \includegraphics[width=\columnwidth]{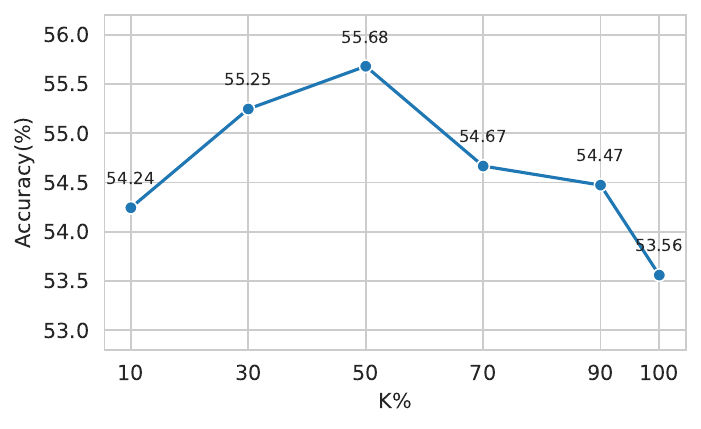}
  \caption{Relationship between average answer accuracy and filter ratio $K\%$.}
  \label{f8}
\end{figure}

We investigate the influence of the filter ratio $K\%$ by experimenting with various values of $K$. We assess the average answer accuracy with $K$ taking values from the set $\{10, 30, 50, 70, 90, 100\}$ using a single iteration, testing on randomly selected 100 questions from each dataset (2,800 questions in total). We maintain a consistent number of training steps across different $K$ values. As illustrated in Figure \ref{f8}, our method achieves optimal results when $K\%$ is set to 50\%.

\subsection{An Example of Generated Stories Alongside the Self-SFT Iterations}

We show an example of the generated stories by the Vicuna model before (iteration 0) and alongside (iteration 1-3) the self-SFT iterations. The commonsense score and BERT similarity both increase along with the iteration, and the answer becomes correct after iteration 2. This further demonstrates the effectiveness of our self-SFT method in generating better stories.

\begin{table*}[!t]
\centering
\small
\resizebox{\linewidth}{!}{
\begin{tabular}{cm{230pt}ccl}
\hline
Iteration & Story & Commonsense Score & BERT Similarity & Predicted Answer \\ \hline
0 & Jane recalls a time when she was in her friend's kitchen and saw her using a glue stick to attach a decoration to a cake. Jane watched as her friend carefully applied the glue to the back of the decoration, then pressed it onto the cake. Jane was impressed by how smoothly the glue stuck to the cake and how easily the decoration stayed in place. She had never thought about adults using glue sticks before, but now she could see how useful they could be for a variety of tasks. & 0.657 & 0.827 & E. kitchen drawer \ding{55} \\ \hline
1 & As a child, Jane used to spend hours crafting with glue sticks. She would often use them to attach paper, cardboard, and other materials to create various projects. She would usually keep a glue stick and a small container of glue near her workspace, which was usually a table in the living room or her bedroom. She would also sometimes use glue sticks to fix small broken objects around the house, such as toys or household items. & 0.692 & 0.830 & C. at school \ding{55} \\ \hline
2 & Jane was working as an intern at a local law firm. One day, her supervisor asked her to help organize the \textbf{office} supplies. As she opened the drawer of the desk she was assigned, Jane noticed a package of glue sticks neatly arranged next to the stapler and scissors. Curiosity piqued, she inquired about the glue sticks. Her supervisor explained that the \textbf{office} frequently used them for various tasks such as attaching documents to presentation boards or assembling reports. Jane was surprised but intrigued by this unconventional use of glue sticks in an \textbf{office} setting. & 0.809 & 0.848 & D. office \ding{51} \\ \hline
3 & I remember one specific experience where I saw adults using glue sticks. It was during my first day of an \textbf{office} internship. I was assigned to help organize some files in the storage room when I noticed a group of adults huddled around a table. Curiosity got the better of me, so I approached them to see what they were doing. To my surprise, they were using glue sticks to attach important documents onto presentation boards. It was fascinating to watch them carefully apply the glue in a controlled manner, ensuring the papers were securely fixed. I realized that adults indeed use glue sticks in an \textbf{office} setting for various tasks. & 0.810 & 0.874 & D. office \ding{51} \\ 
\hline
\end{tabular}}
\caption{\label{t8}
An example of the generated stories of past experiences alongside the iteration of self-SFT. The question is shown in Figure \ref{f0}.
}
\end{table*}

\section{Computational Costs}

For the QA evaluations across the 28 datasets, we use an RTX 3090 GPU. Generating the five stories or rules for a single commonsense question typically takes about 20-30 seconds. Once the stories or rules are generated, answering a question with them as context takes less than a second. Therefore, a complete evaluation of all datasets consumes approximately 170 hours, which spans slightly over a week.

In the iterative self-SFT method employed for fine-tuning the Vicuna model, we use an A100 GPU (80G). The training step in each iteration of the self-SFT method takes around 12 minutes. Additionally, the generation step in each iteration requires about 70 hours to process the 8 datasets.

\section{Licenses of Datasets and Models}

The CommonsenseQA, NumerSense, HellaSWAG, SWAG, and WinoVenti datasets are licensed under the MIT License.
The OpenBookQA, QASC, WinoGrande, and PROST datasets are licensed under the Apache-2.0 license.
The PIQA dataset is licensed under the Academic Free License v. 3.0 license.
The SocialIQA, CommonsenseQA2.0, QuaRel, and QuaRTz datasets are licensed under the CC-BY license.
The SciQ dataset is licensed under the CC-BY-NC-3.0 license.
The ARC and ComVE (Task A) datasets are licensed under the CC-BY-SA license.
The $\alpha$NLI dataset is licensed under the CC-BY-NC-4.0 license.
The COPA dataset is licensed under the BSD 2-Clause license.
The CODAH dataset is licensed under the Open Data Commons Attribution license.
The AI2Sci, Com2Sense, CycIC, and SCT datasets have unknown licenses.

The terms of use for ChatGPT API are in \url{https://openai.com/policies/terms-of-use}. The Alpaca and Vera models are licensed under the MIT License. The Vicuna model is licensed under the Llama 2 Community License Agreement. The BERT model is licensed under the Apache-2.0 license.

\end{document}